%% file: neurips_2023.tex
\documentclass{article}

\PassOptionsToPackage{numbers, sort&compress}{natbib}

\newif\ifarxiv
\arxivtrue

\usepackage[final]{neurips_2023}

\usepackage[utf8]{inputenc} 
\usepackage[T1]{fontenc}    
\usepackage{hyperref}       
\usepackage{xcolor}
\hypersetup{
    colorlinks,
    linkcolor={red!50!black},
    citecolor={blue!80!black},
    urlcolor={blue!80!black}
}
\usepackage{url}            
\usepackage{booktabs}       
\usepackage{amsfonts}       
\usepackage{nicefrac}       
\usepackage{microtype}      
\usepackage{xcolor}         

\usepackage{sidecap}

\usepackage{amsmath, amssymb, url}
\usepackage{bm,xpatch}
\usepackage[capitalize]{cleveref}
\usepackage{xcolor}
\usepackage{fp, tikz, pgfplots}
\usepackage{siunitx}
\usepackage{algorithm}
\usepackage{algpseudocode}
\usepackage{mathtools}
\usepackage{multirow}
\usepackage{booktabs}
\usepackage{interval}
\usepackage{capt-of}
\usepackage{wrapfig}
\usepackage{lipsum}
\usepackage[usestackEOL]{stackengine}
\usepackage{array}
\usepackage{multicol}
\usepackage{wrapfig}
\usetikzlibrary{arrows}
\usetikzlibrary{shapes}
\usetikzlibrary{patterns}
\usetikzlibrary{decorations.pathmorphing}
\usetikzlibrary{decorations.pathreplacing}
\usetikzlibrary{decorations.shapes}
\usetikzlibrary{decorations.text}
\usetikzlibrary{shapes.misc}
\usetikzlibrary{decorations.markings}
\usetikzlibrary{arrows.meta}
\usetikzlibrary{backgrounds}
\usepgfplotslibrary{fillbetween}
\usepgfplotslibrary{statistics}

\usepackage{tikz}
\usepackage{tikz-cd}
\usetikzlibrary{shapes,arrows,fit,backgrounds,positioning}
\usetikzlibrary{decorations.pathmorphing} 
\usetikzlibrary{matrix} 
\usetikzlibrary{arrows} 
\usetikzlibrary{calc} 
\usepackage{verbatim}

\usepackage{amsthm}

\DeclareMathOperator*{\argmin}{arg\,min}

\makeatletter
\def\blfootnote{\gdef\@thefnmark{}\@footnotetext}
\makeatother

\definecolor{tumblue}{HTML}{0065bd} 
\definecolor{tumblue4}{HTML}{98c6ea} 
\definecolor{tumbluered}{HTML}{bd6500}
\definecolor{tumblue4red}{HTML}{eac698}

\newcommand\ko[0]{{\mathcal{K}}}

\newcommand\Set[1]{\mathbb{#1}} 
\newcommand{\tsgn}[1]{{#1}}
\renewcommand{\exp}[1]{\operatorname{e}^{#1}} 

\usepackage{comment}
\usepackage{amsmath}
\usepackage{amssymb}
\usepackage{siunitx}
\usepackage{bm}
\usepackage{multirow}
\usepackage{graphicx}
\usepackage{scalerel}
\usepackage{enumitem}
\usepackage{caption}
\usepackage{subcaption}
\usepackage{todonotes}

\usepackage{palatino}
\let\normalint\int 
\def\int{\displaystyle\normalint} 

\graphicspath{{pics/}}
\newcommand{\Data}{{\Set{D}}}
\newcommand{\trajH}{{{\operatorname{H}}}}

\newcommand{\trajT}{{{T}}}
\newcommand{\datapt}{{{(i)\!}}}
\newcommand{\scale}[1]{{{#1\!}}}
\newcommand{\scalE}[1]{{{#1}}}
\newcommand{\naught}{{_{{0\!}}}}

\renewcommand{\text}[1]{\textnormal{#1}}
\newcommand{\ModOp}{{M}}
\newcommand{\dt}{{\Delta t}}
\newcommand{\eig}{\lambda}

\newcommand{\operator}[1]{\mathcal{#1}}
\newcommand{\raum}[1]{\mathcal{#1}}

\newcommand{\Complex}{\ensuremath{\mathbb{C}}}
\newcommand{\RKHS}{\ensuremath{\raum{H}}}
\newcommand{\dtRKHS}{\ensuremath{\raum{H}^{\scale{\dt}}}}

\newcommand{\Sobo}[2]{\raum{C}}

\newcommand{\ArminKernel}[1][]{Armin-Kernel}
\newcommand{\ArminKernelshort}[1][]{A}

\renewcommand{\d}[1]{\ensuremath{\operatorname{d}\!{#1}}}
\newcommand{\adjoint}[1]{\ensuremath{\left. {#1}\right. ^{\star}}}

\newcommand{\evalat}[2]{\ensuremath{\left. {#1}\right| _{#2}}}
\newcommand{\Erw}{\mathbb{E}}
\newcommand{\Expect}[1]{\mathbb{E}[{#1}]}

\newcommand{\loss}[0]{\mathcal{L}}

\setcitestyle{square,comma, aysep={;}, yysep={,}, notesep={, }}

\newtheorem{definition}{Definition}
\newtheorem{assumption}{Assumption}

\newtheorem{lemma}{Lemma}
\newtheorem{Proof}{Proof}
\newtheorem{theorem}{Theorem}

\newtheorem{proposition}{Proposition}
\newtheorem{remark}{Remark}

\DeclareSymbolFont{myletters}{OML}{ztmcm}{m}{it}
\DeclareMathSymbol{\deig}{\mathord}{myletters}{"15}
\renewcommand{\deig}{\mu}
\usepackage{thmtools,thm-restate}
\newcommand{\addriskplot}[7]{
\addplot[color=green, name path=path1, color=#4, #6] table[x=N,y=y]{#3#1_#2.csv};
\addplot[draw=none, name path=D]
table[x=N,y=ym]{#3#1_#2.csv};
\addplot[draw=none, name path=U]
table[x=N,y=yp]{#3#1_#2.csv};
\addplot[fill=black, fill opacity=0.1, color=#5, #7] fill between [of=D and U];
}

\newcommand{\pathtoresults}[0]{}

\title{Koopman Kernel Regression}

\author{%
  Petar Bevanda \\
TU Munich \\
  \texttt{petar.bevanda@tum.de} \\
  \And
  Max Beier \\
  TU Munich \\
  \texttt{max.beier@tum.de} \\
    \And
  Armin Lederer\\
  TU Munich \\
  \texttt{armin.lederer@tum.de} \\
    \And
  Stefan Sosnowski \\
  TU Munich \\
  \texttt{sosnowski@tum.de} \\
    \And
  Eyke Hüllermeier \\
  LMU Munich \\
  \texttt{eyke@ifi.lmu.de} \\
  \And
  Sandra Hirche \\
  TU Munich \\
  \texttt{hirche@tum.de} \\
}

\begin{document}

\maketitle

\vspace{-\intextsep}
\begin{abstract}

Many machine learning approaches for decision making, such as reinforcement learning, rely on simulators or predictive models to forecast the time-evolution of quantities of interest, e.g., the state of an agent or the reward of a policy. Forecasts of such complex phenomena are commonly described by highly nonlinear dynamical systems, making their use in optimization-based decision-making challenging.
Koopman operator theory offers a beneficial paradigm for addressing this problem by characterizing forecasts via linear time-invariant (LTI) ODEs, turning multi-step forecasts into sparse matrix multiplication.
Though there exists a variety of learning approaches, they usually lack crucial learning-theoretic guarantees, making the behavior of the obtained models with increasing data and dimensionality unclear.
We address the aforementioned by deriving a universal Koopman-invariant reproducing kernel Hilbert space (RKHS) that solely spans transformations into LTI dynamical systems. The resulting \textit{Koopman Kernel Regression (KKR)} framework enables the use of statistical learning tools from function approximation for novel convergence results and generalization error bounds under weaker assumptions than existing work. Our experiments demonstrate superior forecasting performance compared to Koopman operator and sequential data predictors in RKHS.
\end{abstract}
\vspace{-\intextsep}
\section{Introduction}

    Dynamical systems theory is a fundamental paradigm for understanding and modeling the time evolution of a phenomenon governed by certain underlying laws. Such a perspective has been successful in describing countless real-world phenomena, ranging from engineering mechanics \cite{Meriam2020EngineeringDynamics} and human movement modeling \cite{Billard2022LearningApproach} to molecular and quantum systems \cite{May2011ChargeSystems,Johansson2012QuTiP:Systems}.
However, as the laws governing dynamical systems are often unknown, modeling and understanding the underlying phenomena may have to rely on data rather than first principles. 
    In this regard, machine learning methods, which have shown immense potential in tackling complex tasks in domains such as language models \cite{Brown2020LanguageLearners} and computer vision \cite{Radford2021LearningSupervision}, are coming to the fore.
Though powerful, state-of-the-art neural vector fields \cite{Chen2018NeuralEquations} or flows \cite{Bilos2021NeuralODEs} commonly compose highly nonlinear maps for forecast, i.e. computing
\vspace{-0.33\intextsep}
\begin{equation}\label{ODESolve}
    {x}\left(t\right)={x}(0) +\normalint_{0}^{t} f({x}(t)) \mathrm{d} t
\end{equation} for, e.g. a scalar ODE $\dot{x}=f(x)$.
Hence, it is often challenging to use such models in optimization-based decision making that relies on simulators or predictive models, e.g., reinforcement learning \cite{Janner2021OfflineProblem,Chua2018DeepModels,Theodorou2010ALearning}.
A particularly beneficial perspective for dealing with the aforementioned comes from Koopman operator theory \cite{Budisic2012AppliedKoopmanism,KoopBook,Mezic2004ComparisonBehavior,Mezic2005SpectralDecompositions}. Through a point-spectral decomposition of Koopman operators, forecasts become superpositions of solution curves of a set of linear ODEs $\{\dot{z}_j=\lambda_jz_j\}_{j=1}^D$
\vspace{-0.4\intextsep}
\begin{equation}\label{KoopSolve}
    {x}\left(t\right)=\sum_{k=1}^D \operatorname{e}^{\lambda_j t}z_j(0), \qquad \{x \overset{\text{lift}~g_j}{\longmapsto
    } z_j\}_{j=1}^{D} 
\end{equation}
where a vector valued function $\operatorname{span}(\{g_j\}_{j=1}^{D})$ ``lifts'' $x$ onto a manifold $\Set{Z} :=\operatorname{span}\left(
\{z_j\}_{j=1}^D\right)$. Throughout, we refer to these models as \textit{linear time-invariant (LTI) predictors}.
The learning objective of such representations is twofold: spanning system trajectories by the learned manifold $\Set{Z}$ and constraining the LTI dynamics to it. The latter is a long-standing challenge of Koopmanism \cite{Brunton2016KoopmanControl,Brunton2019Data-DrivenEngineering,Otto2021AnnualSystems,Bevanda2021KoopmanControl,Brunton2022ModernSystems}, as manifold dynamics of existing approaches ``leak-out'' \cite{Mezic2020SpectrumGeometry} and limit predictive performance.
 \begin{wrapfigure}[]{r}{0.66\textwidth}
\vspace{-\intextsep}
  \begin{center}
    \includegraphics[width=0.6\textwidth]{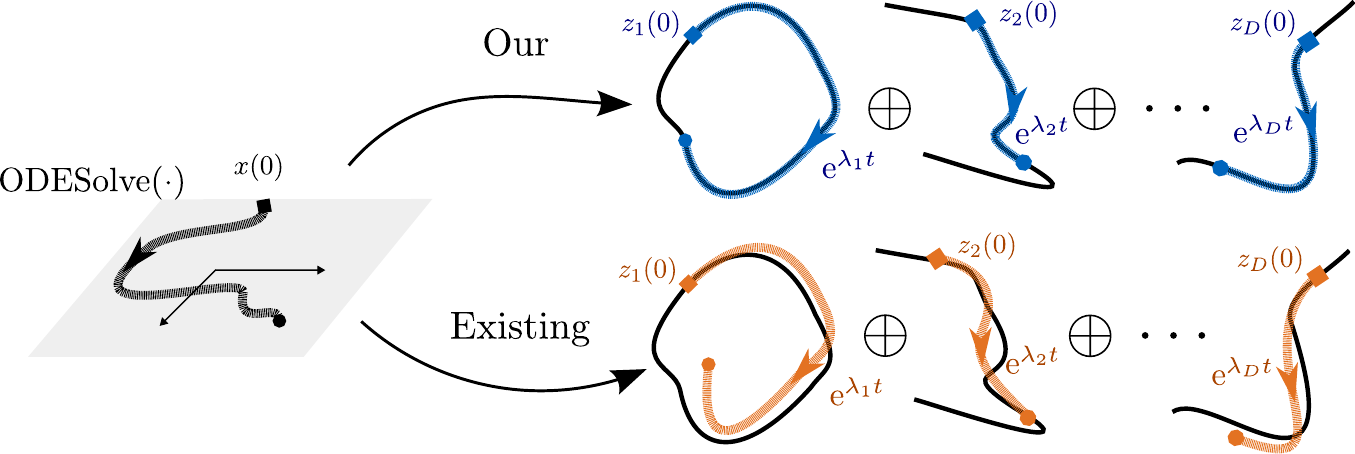}
  \end{center}
  \vspace{-0.1cm}
  \caption{Illustration of on-manifold dynamics of LTI predictors.}
  \vspace{-0.5cm}
  \label{fig:manifold}
\end{wrapfigure}
To tackle the aforesaid, we connect the representation theories of reproducing kernel Hilbert spaces (RKHS) and Koopman operators. As a first in the literature, we derive a universal kernel whose RKHS exclusively spans manifolds invariant under the dynamics, as depicted in Figure \ref{fig:manifold}.
A key corollary of unconstrained manifold dynamics is the lack of essential learning-theoretic guarantees, making the behavior of existing learned models unclear for increasing data and dimensionality.
To address this, we utilize equivalences to function regression in RKHS to formalize a statistical learning framework for learning LTI predictors from sample trajectories of a dynamical system. This, in turn, enables the use of
statistical learning tools from function approximation for novel convergence results
and generalization error bounds under weaker assumptions than before \cite{Klus2020EigendecompositionsSpaces,Kostic2022LearningSpaces,Kostic2023KoopmanEigenvalues}. Thus, we believe that our Koopman Kernel Regression (KKR) framework takes the best of both RKHS and Koopmanism by leveraging modular kernel learning tools to build provably effective LTI predictors.

The remainder of this paper is structured as follows: We briefly introduce LTI predictors
and discuss related work in Section \ref{sec:probStat}. The derivation of the KKR framework, including the novel Koopman RKHS, is presented in Section \ref{sec:KKR}. In Section \ref{section:lernGaran},
we show the novel learning guarantees in terms of convergence and generalization error bounds. They are validated in comparison to the state-of-the-art through numerical experiments in Section \ref{sec:numExp}.

{\bf {Notation}~} Lower/upper case bold symbols $\bm{x}/\bm{X}$ denote spatial vector/matrix-valued quantities.
    A \textit{trajectory} defines a curve $\bm{x}_{\trajT}\subset\mathbb{X}$ traced out by the flow over time $\Set{T}\tsgn{=}\left[0, T\right]$ from any $(\tau,\bm{x}) \in \Set{T} {\times} \Set{X}$. In discretizing $\Set{T}$, collection of points $\bm{x}_{\trajH}\subset \mathbb{X}$ from discrete time steps $\Set{H} {=}\{t\naught \cdots t_\trajH\}$ is considered. The state/output trajectory spaces are denoted as $\Set{X}_{\trajT}\tsgn{\subseteq }L^2(\Set{T}, \Set{X})$ / $\Set{Y}_{\trajT}\tsgn{\subseteq }L^2(\Set{T},\Set {Y})$, with discrete-time analogues $\Set{X}_{\trajH}\tsgn{\subseteq }\ell^2(\Set{H},\Set{X})$ / $\Set{Y}_\trajH\tsgn{\subseteq}\ell^2(\Set{H},\Set {Y})$ with domain and co-domain separated by ``$,$''. The vector space of continuous functions on $\Set{X}_\trajT$ endowed with the topology of uniform convergence on compact domain subsets is denoted $C(\Set{X}_\trajT)$. The collection of {bounded} linear operators from $\Set{Y}_\trajT$ to $\Set{Y}_\trajT$ is denoted as $\mathcal{B}(\Set{Y}_\trajT)$. The adjoint of $\mathcal{A} \in \mathcal{B}(\cdot)$ is $\mathcal{A}^*$. Discrete-time eigenvalues read $\mu{:=}\operatorname{e}^{\eig \Delta t}, \eig \in \Set{C}$. A random variable $X$ defined on a probability space $(\Omega, \mathcal{A}, \rho)$ has expectation $\Expect{X}= \normalint_{\Omega}X(\omega)\rho(\omega)$.
     \vspace{-0.25\intextsep}
\section{Problem Statement and Related Work}\label{sec:probStat}
 \vspace{-0.5\intextsep}
To begin, we formalize our problem statement and put our work into into context with existing work.
 \vspace{-0.25\intextsep}
\subsection{{Problem Statement}}
 \vspace{-0.25\intextsep}
Consider a forward-complete system\footnote{Although we outline the scalar output case for ease of exposition, expanding to a vector-valued case is possible w.l.o.g. If required, forward completeness can be relaxed to unboundedness observability \cite{Angeli1999ForwardCharacterizations,Andrieu2006OnObserver}. 
}  comprising a {nonlinear} state-space model
\begin{subequations}\label{eq:nonlinearPredictor}
    \begin{align}
            \dot{\bm{x}} &= \bm{f}(\bm{x}), ~\bm{x}\naught=\bm{x}(0) \label{eq:NLTI},\\
            {y}&= q(\bm{x}) \label{eq:Nrecon},
    \end{align}
\end{subequations}
on a compact domain $\Set{X} \subset \Set{R}^{d}$ with a {quantity of interest} ${q}\!\in\! C(\Set{X})$.
The above system class includes all systems with Lipschitz flow $\bm{F}^{t}(\bm{x}{\naught}):=\normalint_0^{t} \bm{f}(\bm{x}(\tau)) d\tau$, e.g., mechanical systems \cite{Krstic2009Forward-CompleteSystems}.

Inspired by the spectral decomposition of Koopman operators, we look to replace the nonlinear state-space model \eqref{eq:nonlinearPredictor} by an \textit{LTI predictor}
\begin{subequations}\label{eq:GenlinearPredictor}
    \begin{align}
            \dot{\bm{z}} &= \bm{A}\bm{z},  ~\bm{z}\naught=\bm{g}(\bm{x}\naught), \label{eq:LTI}\\
            {{y}}&= \bm{c}^{\top}\!\bm{z}   \label{eq:recon},
    \end{align}
\end{subequations} 
\looseness=-1
with $\bar{D}\in \Set{N}$ and $\bm{g}$ a $\bar{D}$-dimensional function approximator dense in $C(\Set{X})$. Then, from initial conditions $\Set{X}\naught \subseteq \Set{X}$ that form a \textit{non-recurrent} domain $\Set{X}_\trajT$, \eqref{eq:GenlinearPredictor} admits a universal approximation of the flow of \eqref{eq:nonlinearPredictor} such that $\forall \varepsilon > 0,~\exists \bar{D}$ so that $\operatorname{sup}_{\bm{x} \in \Set{X}\naught}|y_{\trajT}(t) - \bm{c}^{\top}\!\operatorname{e}^{\bm{A}t}\bm{g}(\bm{x})| < \varepsilon, ~ \forall t \in [0,T]$  \cite{Korda2020OptimalControl}\footnote{Background on prerequisite Koopman operator theory can be found in the supplementary material.}.
In this work, we aim to find a solution to the following constrained, functional optimization problem
\begin{subequations}\label{prob:intract}
\vspace{-.15truecm}
\begin{enumerate}[label={\rm \textbf{(OR)}},leftmargin=7ex,nolistsep]
\item\label{eq:lossRecon} \emph{Output reconstruction:} 
\vspace{-.25truecm}
\begin{align}
   & \min_{\bm{c},\bm{g},\bm{A}}\|{y}_\trajT- \bm{c}^{\top}\!\bm{g}_{\trajT}\|_{\Set{Y}_\trajT}, \label{prob:obj}\end{align}
   \end{enumerate}
\vspace{-.25truecm}
\begin{enumerate}[label={\rm \textbf{(KI)}},leftmargin=7ex,nolistsep]
\item\label{eq:KIconst} \emph{Koopman-invariance:}
\vspace{-.25truecm}
\begin{align} \qquad \qquad \text{such that} & \quad \bm{g}(\bm{x}(t)) = \operatorname{e}^{\bm{A}t}\bm{g}(\bm{x}(0)), \quad ~~\forall t \in [0,T] \label{prob:const}. 
\end{align}
\end{enumerate}
\end{subequations}
Although the sought-out model \eqref{eq:GenlinearPredictor} is simple, the above problem is non-trivial and much of the existing body of work utilizes different simplifications that often lead to undesirable properties. In the following, we elaborate on these properties and motivate our novel sample-based solution to \eqref{prob:intract}, which remains relatively simple but nonetheless ensures a well-defined solution with strong learning guarantees.

\subsection{Related Work}
{\bf Koopman operator regression in RKHS~}
Equipped with a rich set of estimators, operator regression in RKHS seeks a sampled-data solution to
\begin{align}
   & \min_{\bm{A}}\|\bm{g}(\bm{x}(t))- \bm{A}^{*}\!\bm{g}(\bm{x}(0))\|_{L^2}, \label{prob:KOR}
\end{align}
\looseness=-1
with $\bm{A}$ a Hilbert-Schmid operator \cite{Klus2020eig} --- commonly known as KRR, and EDMD (PCR) or RRR when under different fixed-rank constraints \cite{Kostic2022LearningSpaces, Kostic2023KoopmanEigenvalues}. The choice of RKHS $\bm{g}$ is commonly one that is dense in a suitable $L^2$ space, e.g. that of the RBF kernel.
By an additional projection, a quantity of interest can be predicted via a mode decomposition of the estimated operator, leading to a model akin to \eqref{eq:GenlinearPredictor}. 
In the light of \eqref{prob:intract}, the feature map $\bm{g}$ is predetermined while violating \ref{eq:KIconst} is merely minimized for a single time-instant $t$. As a consequence, such approaches are oblivious to the time-series structure --- offering limited predictive power over the time interval $[0,T]$ of a trajectory as displayed in Figure \ref{fig:manifold}. 
The extent to which \ref{eq:KIconst} is violated due to spectral properties \cite{ColbrookResidualKoopmanism} or estimator bias \cite{Kostic2023KoopmanEigenvalues} is known as spectral pollution \cite{Lewin2010SpectralIt}. The strong implications of this phenomena, motivate regularization \cite{Khosravi2022} and spectral bias measures \cite{Kostic2023KoopmanEigenvalues} to reduce its effects.
Due to the above challenges, guarantees for Koopman operator regression (KOR) have only recently gained increased attention. Often, however, existing theoretical results \cite{Klus2020eig,Philipp2023ErrorOperator} are generally not applicable to nonlinear dynamics \cite{Das2020} due to the commonly unavoidable misspecification \cite{Valva2023ConsistentCompactification} of the problem \eqref{prob:KOR} incurred by neglecting \ref{eq:KIconst}.
The first more general statistical learning results \cite{Kostic2022LearningSpaces,Kostic2023KoopmanEigenvalues} are derived in a stochastic setting under the assumption that the underlying operator is compact and self-adjoint. In stark contrast, the same set of assumptions is restrictive for the deterministic setting \cite{Valva2023ConsistentCompactification}: compactness only holds for affine deterministic dynamics \cite{Singh1979CompactOperators,Ikeda2022BoundednessFunctions} while self-adjointness is known to generally not hold for Koopman operators \cite{Cvitanovic2016Chaos:Quantum,KoopBook,Mezic2022OnOperator}. Regardless of the setting, however, the state-of-the-art exhibits alarming properties: forecasting error not necessarily vanishing with LTI predictor \eqref{eq:GenlinearPredictor} rank \cite[Theorem 1]{Kostic2022LearningSpaces} and risk based on a single time-instant.

{\bf Learning via Koopman eigenspaces~} 
Geared towards LTI predictors and closer to our own problem setting \eqref{sec:probStat}, another distinct family of approaches aims to directly learn the operator's invariant subspaces \cite{Bevanda2022LearningApproach,Bevanda2022DiffeomorphicallyOperators,Bollt2021GeometricRepresentation,Korda2020OptimalControl}. The goal is to fit $\bm{g}(\cdot)$ based on approximate Koopman operator eigenfunctions that still fit the output of interest \ref{eq:lossRecon}. 
However, existing data-driven approaches in this line of work rely on ad-hoc choices and lack essential learning-theoretic properties such as feasibility and uniqueness of solutions --- prohibiting provably accurate and automated LTI predictor learning.

{\bf Kernels for sequential data~}
Motivated by the lack of priors that naturally incorporate streaming and sequential data, there is an increasing interest in \textit{signatures} \cite{Kiraly2016KernelsData}. They draw from the rich theory of controlled differential equations (CDEs) \cite{Lyons1998DifferentialSignals,Friz2010MultidimensionalPaths} and build models that depend on a time-varying observation history. An RKHS suitable for sequence modeling is induced by a signature transformation of a base/static RKHS. Generally, if the latter is
universal, so are the signature kernels \cite{Lee2023TheKernel}. While arguably more general and well-versed for discriminative and generative tasks \cite{Lemercier2021SigGPDE:Data}, forecasting using signature kernels \cite{Salvi2021ThePDE} comes at a price, as their nonlinear dependence on observation streams leads to a significant complexity increase compared to LTI predictors.
\looseness=-1

Motivated by the restrictions of existing Koopman-based predictors, we propose a \textit{function approximation} approach that exploits exploits time-series data and Koopman operator theory to provably learn LTI predictors. Through a novel \textit{invariance transform} we can satisfy \ref{eq:KIconst} by construction and directly minimize the forecasting risk over an entire time-interval \ref{eq:lossRecon}. In simple terms: Koopman operator regression fixes $\bm{g}(\cdot)$ and regresses $\bm{A}$ and $\bm{c}$ in \eqref{prob:intract}, whereas our KKR approach selects $\bm{A}$ to jointly regress $\bm{c}$ and $\bm{g}(\cdot)$. Similar in spirit to generalized Laplace analysis \cite{Mezic2013AnalysisOperator, Mezic2020SpectrumGeometry}, our approach allows the construction of eigenmodes from data without inferring the operator itself. Crucially, we demonstrate that selecting $\bm{A}$ requires no prior knowledge as confirmed by our theoretical results and experiments.
To facilitate learning LTI predictors, we derive \textit{universal} RKHSs that are \textit{guaranteed} to satisfy \ref{eq:KIconst} over trajectories --- a first in the literature. The resulting equivalences to function regression in RKHS allow for more general and complete learning guarantees in terms of consistency and risk bounds that are free of restrictive operator-theoretic assumptions.
\section{Koopman Kernel Regression}\label{sec:KKR}
\label{section:KKR}

With the optimization \eqref{prob:intract} being prohibitively hard due to nonlinear and possibly high dimensional constraints, we eliminate the constraints \eqref{prob:const} by enforcing the feature map $\bm{g}(\cdot)$ to have the dynamics of intrinsic LTI coordinates associated with Koopman operators, i.e., their (open) eigenfunctions \cite{Mezic2020SpectrumGeometry}.
 \begin{definition}\label{def:KEIGS}
	A Koopman eigenfunction $\phi_{\eig {\in} \Set{C}} {\in} C(\Set{X})$ satisfies $\phi_{\eig}(\bm{x})\tsgn{=}\operatorname{e}^{-\eig t}\phi_{\eig}(\bm{F}^{t}(\bm{x})), \forall t \in [0,T]$.
\end{definition}
It is proven that Koopman eigenfunctions from Definition \ref{def:KEIGS} are universal approximators of continuous functions \cite{Korda2020OptimalControl} --- making them a viable replacement for the feature map $\bm{g}(\cdot)$ in \eqref{eq:GenlinearPredictor}. However, following their definition, it is evident that Koopman eigenfunctions are by no means arbitrary due to their inherent dependence on the dynamics' flow.
Using the well-established fact that Koopman operators compose a function with the flow, i.e., $\ko^t{g}(\cdot) = {g}(\bm{F}^t(\cdot))$, it becomes evident the eigenfunctions from Definition \ref{def:KEIGS} are (semi)group invariants, as they remain 
unchanged after applying $\{\operatorname{e}^{-\eig t}\ko^{t}\}^{T}_{t\tsgn{=}0}$.
Thus, inspired by the seminal work of Hurwitz on constructing invariants \cite{Hurwitz1897UberIntegration}, we can equivalently reformulate \eqref{prob:intract} as an unconstrained problem and jointly optimize over eigenfunctions\footnote{Proofs for all theoretical results can be found in the supplementary material.
}.

\begin{restatable}[Invariance transform]{lemma}{InvarinceTransform}\label{lemma:Invariants}
Consider a function $g\in C(\Set{X}\naught)$ over a set of initial conditions $\Set{X}\naught \subseteq \Set{X}$ that form a \textit{non-recurrent} domain $\Set{X}_\trajT$. The \text{invariance transform} $\mathcal{I}^T_{\eig}$ transforms $g$ into an Koopman eigenfunction $\phi_{\eig} \in C(\Set{X}_\trajT)$ for \eqref{eq:NLTI} with LTI dynamics described by $\eig \in \Set{C}$
\begin{equation}\label{eq:EFtransform}
    \phi_{\eig}(\bm{x}_\trajT) = \mathcal{I}^T_{\eig}{g}(\bm{x}\naught) :=  \int^{T}_{\tau=0}\operatorname{e}^{-\eig(\tau - t)} g(\bm{F}^{\tau}(\bm{x}\naught))d \tau.
\end{equation}
\end{restatable}
The above Lemma \ref{lemma:Invariants} is a key stepping stone towards deriving a representer theorem for LTI predictors. However, it is also interesting in its own right as it provides an explicit expression for the flow of an eigenfunction from any point in the state space.
Thus, it provides a recipe to obtain a function space that fulfills~\ref{eq:KIconst} by construction. 
As we show in the following, a sufficiently rich set of eigenvalues \cite{Korda2020OptimalControl} and Lemma \ref{lemma:Invariants} will allow for a reformulation of \eqref{prob:intract} into an unconstrained problem
\begin{equation}\label{prob:UnconIntrac} 
  \min_{\ModOp}\|{y}_\trajT- \ModOp(\bm{x}_\trajT)\|_{\Set{Y}_\trajT}.
\end{equation}
where the operator $\ModOp(\cdot) {:=} \bm{1}^\top [\phi_{{\eig}_{1}}(\cdot) \cdots \phi_{{\eig}_{\bar{D}}}(\cdot)]^{\top}$ is universal and consisting of Koopman-invariant functions.
\subsection{Functional Regression Problem}\label{subsec:functional}%
Notice that the problem reformulation \eqref{prob:UnconIntrac} is still intractable, as a closed-form expression for the flow map is generally unavailable even for known ODEs. This requires integration schemes that can introduce inaccuracies over a time interval $[0,T]$. Thus, to make the above optimization problem tractable, data samples are used --- ubiquitous in learning dynamical systems. 
\begin{assumption}\label{ass:dataT}
A collection of $N$ pairs of trajectories
$\Data_N\tsgn{=}\{\bm{x}_{\trajT}^\datapt,{y}_{\trajT}^\datapt\}^N_{i=1} \!\in\! (\Set{X}_{\trajT} \tsgn{\times} \Set{Y}_{\trajT})^{N}$ is available.\looseness=-1
\end{assumption}
    By aggregating different invariance transformations \eqref{eq:EFtransform} into the \textit{mode decomposition operator} \begin{equation}\label{eq:ModeOp}
        {\ModOp}(\cdot) \tsgn{\equiv} \textstyle{\sum_{j=1}^{\bar{D}}}\phi_{{\eig}_{j}}(\cdot)\!\!:~\Set{X}_{\trajT} \mapsto \Set{Y}_{\trajT},
    \end{equation} we can formulate a supervised learning approach in the following.
    
\phantomsection{\bf Learning Problem~}\label{eq:preRKHS} With Assumption \ref{ass:dataT} and Lemma \ref{lemma:Invariants}, the sample-based approximation of problem \eqref{prob:UnconIntrac} reduces to solving 
\begin{equation}
\min_{{\ModOp}}\textstyle\sum^{N}_{i=1}\|{y}^\datapt_{\trajT}-{\ModOp}(\bm{x}^\datapt_{\trajT})\|_{\Set{Y}_{\trajT}}. 
\end{equation}
while preserving the mode decomposition structure \eqref{eq:ModeOp}.
To realize the above learning problem, we resort to the theory of {reproducing kernels} \cite{Scholkopf2018LearningKernels,IngoSteinwart2008SupportMachines} and look for an operator $\hat{\ModOp} \in \RKHS$, where $\RKHS$ is an RKHS. A well-established approach using RKHS theory is to select $\hat{\ModOp}$ as a solution to the \textit{regularized least squares problem}
\begin{equation}\label{eq:LSreg}
\hat{\ModOp} = \argmin_{M \in \RKHS}\sum^{N}_{i=1}\|{y}^\datapt_{\trajT}-{\ModOp}(\bm{x}^\datapt_{\trajT})\|^2_{\Set{Y}_{\trajT}}+\gamma\|{{\ModOp}}\|^2_{\RKHS},
\end{equation}
with $\gamma\! \in \!\Set{R}_{+}$ and $\|\tsgn{\cdot}\|_{\RKHS}$ a corresponding RKHS norm. As our target is a function-valued mapping $M(\cdot)$ -- an operator -- $\|\tsgn{\cdot}\|_{\RKHS}$ is induced by an {\textit{operator}-valued} kernel ${K}\!:\Set{X}_{\trajT}\tsgn{\times}\Set{X}_{\trajT}\!\mapsto \mathcal{B}(\Set{Y}_{\trajT})$ mapping to the space of bounded operators over the output space \cite{Kadri2016Operator-valuedData}. The salient feature of the above formulation \eqref{eq:LSreg} is its well-posedness: its solution exists and is unique for any $\RKHS$, expressed as \begin{equation}
    \hat{\ModOp}(\cdot)\tsgn{=}\textstyle\sum_{i=1}^N  {K}(\cdot, \bm{x}_{\trajT}^\datapt) {\beta}_i, \quad \beta_i \in \Set{Y}_{\trajT}
\end{equation} through a representer theorem \cite{Micchelli2005OnFunctions}.
 Still, due to the Koopman-invariant structure \eqref{eq:ModeOp} from Lemma \ref{lemma:Invariants}, the choice of the RKHS $\RKHS$ for $\hat{\ModOp}$ is not arbitrary. Thus, the question is how to craft $\RKHS$ so the solution $\hat{\ModOp}$ is decomposable into Koopman operator eigenfunctions \eqref{eq:ModeOp}, forming an \textit{LTI predictor}.

Firstly, it is obvious that \eqref{eq:ModeOp} consists of summands that may lie in different RKHS, denoted as $\{\RKHS^{\eig_j}\}_{j=1}^{\bar{D}}$. Then, $\RKHS$ is constructed from the following direct sum of Hilbert spaces \cite{Aronszajn1950TheoryKernels}:
 \begin{equation}\label{prodTOp}
    \tilde{\RKHS} = {\RKHS^{\eig_1} \oplus\cdots \oplus\RKHS^{\eig_{\bar{D}}}} ~~ \operatorname{so~that} ~~ \mathcal{H} = \operatorname{range}(\mathcal{S}){:=} \{ f_{1}+\ldots+f_{\bar{D}}: f_{1} \tsgn{\in} \RKHS^{\eig_1},\ldots,f_{\bar{D}}\tsgn{\in}\RKHS^{\eig_{\bar{D}}} \}
\end{equation}
with $\mathcal{S}\!: \tilde{\RKHS} \rightarrow \RKHS, (f_{1}\cdots f_{\bar{D}}) \mapsto f_{1}+\ldots+f_{\bar{D}}$  the summation operator \cite{Hotz2012RepresentationKernels}.
Thus, to construct $\RKHS$, a specification of the RKHS collection $\{\RKHS^{\eig_j}\}_{j=1}^{\bar{D}}$ is required, so that it represents Koopman eigenfunctions from \eqref{eq:ModeOp}.
 \begin{theorem}[Koopman eigenfunction kernel]\label{thm:eigKern}
    Consider trajectory data $\{\bm{x}_{\trajT}^\datapt\}^{N}_{i=1}$ from Assumption \ref{ass:dataT}, a $\eig \in \Set{C}$ and a universal (base) kernel $k\!:\Set{X} \tsgn{\times} \Set{X} \mapsto \Set{R}$. Then, the kernel $K^{{\eig}}:\Set{X}_{\trajT} \tsgn{\times} \Set{X}_{\trajT} \mapsto \mathcal{B}(\Set{Y}_{\trajT})$
\begin{equation}\label{eq:KEIGKern}
        K^{{\eig}}(\bm{x}_{\trajT},\bm{x}_{\trajT}^{\prime}) =\int^T_{\tau=0} \int^T_{{\tau}^{\prime}=0} \operatorname{e}^{-\eig ( \tau - t)} k\left(\bm{x}_{\trajT}^{}(\tau),\bm{x}_{\trajT}^{\prime}(\tau^{\prime})\right)\operatorname{e}^{-\eig^*(\tau^{\prime}-t)} d \tau d\tau^{\prime},
\end{equation}
\begin{enumerate}[label=(\roman*),leftmargin=*]
\vspace{-.45truecm}
    \item  defines an RKHS $\RKHS^{\eig}$,
    \vspace{-.25truecm}
    \item is universal for every eigenfunction of Definition \eqref{def:KEIGS} corresponding to $\lambda$,
    \vspace{-.25truecm}
    \item induces a data-dependent function space
    $\operatorname{span} \big\{K^{\eig}(\cdot, \bm{x}^\scalE{(1)}_\trajT), \ldots, K^{\eig}(\cdot, \bm{x}^\scalE{(N)}_\trajT) \big\}$ that is Koopman-invariant over trajectory-data $\textstyle\{\bm{x}_{\trajT}^\datapt\}^{N}_{i=1}$.
    \end{enumerate}
 \end{theorem}
In Theorem \ref{thm:eigKern}, we derive an eigenfunction RKHS by defining its corresponding kernel that embeds the invariance transformation \eqref{eq:EFtransform} over data samples. Also, we would like to highlight that the above result addresses a long-standing open challenge in the Koopman operator community \cite{Mezic2020SpectrumGeometry,Brunton2022ModernSystems,Bevanda2021KoopmanControl}, i.e., defining universal function spaces that are guaranteed to be Koopman-invariant.
 Now, we are ready to introduce the \textit{Koopman kernel} as the kernel obtained by combining ``eigen-RKHS'' as described in \eqref{prodTOp}.\looseness=-1
\begin{proposition}
 [Koopman kernel]\label{prop:koKern}
    Consider trajectory data $\Data_N$ of Assumption \ref{ass:dataT} and a set of kernels $\{K^{{\eig}_j}\}^{\bar{D}}_{j=0}$ from Theorem \ref{thm:eigKern}. Then, the kernel $K:\Set{X}_{\trajT} \tsgn{\times} \Set{X}_{\trajT} \mapsto \mathcal{B}(\Set{Y}_{\trajT})$ given by 
\begin{equation}\label{eq:KoKern}
        K(\bm{x}_{\trajT},\bm{x}_{\trajT}^{\prime}) = \textstyle\sum^{\bar{D}}_{j=1} K^{\eig_j}(\bm{x}_{\trajT},\bm{x}_{\trajT}^{\prime})
\end{equation}
\begin{enumerate}[label=(\roman*),leftmargin=*]
\vspace{-.35truecm}
    \item defines an RKHS $\RKHS:=\mathcal{S}(\RKHS^{{\eig}_1} \oplus \cdots \oplus \RKHS^{{\eig}_{\bar{D}}})$,
    \vspace{-.25truecm}
    \item is universal for any output \eqref{eq:Nrecon}, provided a sufficient amount\footnote{Sufficient amount is a rich enough set of eigenvalues $\{\operatorname{e}^{\eig_j [0,T]}\}^{\bar{D}}_{j=1}$ from $\overline{\Set{B}_1(\bm{0})}$ in $\Set{C}$ \cite[Theorem 3.0.2]{Kuster2015TheSystems}.}
    of eigenspaces $\bar{D}$.
\end{enumerate}
\end{proposition}
Above, we have derived the ``Koopman-RKHS'' $\RKHS$ for solving the problem \eqref{eq:LSreg} with a universal RKHS spanning Koopman eigenfunctions. 
 Thus, the sample-data solution for an eigenfunction flow follows from the functional regression problem \eqref{eq:LSreg} and takes the form $\phi_{\eig_j}(\cdot)= \textstyle\sum_{i=1}^N  {K}^{\eig_j} (\cdot, \bm{x}_{\trajT}^\datapt) {\beta}_i,~\beta_i \! \in\! \Set{Y}_{\trajT}$ --- providing a basis for the LTI predictor.
\subsection{Practicable LTI Predictor Regression}\label{subs:VVkern}
As a functional approximation problem, the solution of \eqref{eq:LSreg} is not parameterized by vector-valued coefficients, but rather functions of time. Although there are a few options to deal with function-valued solutions \cite{Kadri2016Operator-valuedData}, we consider a vector-valued solution. A common drawback of such a discretization involves the loss of the inter-sample relations along the continuous signal. Crucially, this problem does not apply in our case, as the inter-sample relationships remain modeled for the discrete-time ``Koopman kernel'' due to its causal structure. Importantly,
the vector-valued solution allows us to preserve all of the desirable properties derived in the continuous case.

Consider sampling $[0,T]$ at $H\tsgn{=}T/\dt$ regular intervals to yield a discrete-time dataset from Assumption \ref{ass:dataT}, discretized at points $\Set{H} \equiv\{t\naught\cdots\!t_H\}$. As a discretization of a function over time, with a slight abuse of notation, we denote the target vectors as $y_{\trajH} = [y(t\naught) \cdots y(t_\trajH)]^{\top}$. Thus, we are solving the time- and data-discretized version of the problem \eqref{prob:intract} that takes the form of a linear coregionalization model \cite{Alvarez2011KernelsReview,Lederer2021TheControl}.
\begin{restatable}[Time-discrete {Koopman kernel}]{corollary}{DTkoopKern}\label{cor:trajK}
Consider trajectory data $\{\bm{x}_{\trajH}^\datapt\}^{N}_{i=1}$ and let ${\deig_j}\tsgn{:=}\operatorname{e}^{\eig_j \dt}$, $\bm{\deig}^{\top}_j \tsgn{:=} [\deig_j^0 \cdots \deig_j^{H}]$. Then, the scalar-induced matrix kernel $\bm{K}^{{\deig}_j}\!\!:\Set{X}_{\trajH}  \tsgn{\times}\Set{X}_{\trajH}  \mapsto \mathcal{B}(\Set{Y}_{\trajH})$
\vspace{-.0truecm}
\begin{equation}\label{eq:KEIGKernTrajDT}
\bm{K}^{{\deig}_j}(\bm{x}_{\trajH},{\bm{x}_{\trajH}}^\prime) = \bm{\deig}_j {\bm{\deig}^*_j}^{\top} \underbrace{\textstyle{\frac{1}{(H\tsgn{+}1)^2}}\textstyle{\sum^H_{m=0} \sum^H_{n=0}} \deig_j^{\scaleto{-m}{5pt}} k^j\left(\bm{x}_{\trajH}(t_m),{\bm{x}_{\trajH}}^\prime(t_n)\right){\deig}_j^{*\scaleto{-n}{5pt}}}_{k^{\deig_j} \!(\bm{x}_{\trajH},{\bm{x}_{\trajH}}^\prime)},
\end{equation} 
satisfies the properties (i)--(iii) from Theorem \ref{thm:eigKern} over $\Set{H}$, so that it defines an RKHS $\RKHS^{\deig_j}$, is universal per Definition \ref{def:KEIGS} over $\Set{H}$ with $\operatorname{span}\{\bm{K}^{\deig}(\cdot, \bm{x}^\scalE{(1)}_\trajH), \ldots, \bm{K}^{\deig}(\cdot, \bm{x}^\scalE{(N)}_\trajH)\}$ \ref{eq:KIconst} over $\textstyle\{\bm{x}_{\trajH}^\datapt\}^{N}_{i=1}$.
Given a collection of kernels $\{\bm{K}^{{\deig}_j}\}^{\bar{D}}_{j=0}$, the matrix \text{Koopman kernel} $\bm{K}^{{\deig}_j}\!\!:\Set{X}_{\trajH}  \tsgn{\times}\Set{X}_{\trajH} \mapsto \mathcal{B}(\Set{Y}_{\trajH})$
\vspace{-.0truecm}
\begin{equation}\label{eq:KEIGKernTrajDT2}
\bm{K}(\bm{x}_{\trajH},{\bm{x}_{\trajH}}^\prime) =  \textstyle\sum^{\bar{D}}_{j=1} \bm{K}^{\deig_j}(\bm{x}_{\trajH},{\bm{x}_{\trajH}}^\prime),
\end{equation}
satisfies the properties (i)--(ii) from Proposition \ref{prop:koKern} over $\Set{H}$, defining RKHS $\dtRKHS\!:=\!\mathcal{S}(\RKHS^{{\deig}_1} \!\oplus\cdots \oplus\! \RKHS^{{\deig}_{\bar{D}}})$.
\end{restatable}
Now, we are fully equipped to obtain the time-discrete solution to our initial problem \eqref{prob:intract} provided a dataset of trajectories.
Before presenting the solution to Koopman Kernel Regression, we introduce some helpful shorthand notation.
 We use the following kernel matrix abbreviations: $k_{\bm{X}\bm{X}} = [k(\bm{x}^{(a)},\bm{x}^{(b)})]^N_{a,b=1}$, $k({\bm{x},\bm{X}}) = [k(\bm{x},\bm{x}^{(b)})]^N_{b=1}$, $\bm{K}_{\bm{X}\bm{X}} = [\bm{K}(\bm{x}^{(a)},\bm{x}^{(b)})]^N_{a,b=1}$ and $\bm{K}({\bm{x},\bm{X}}) = [\bm{K}(\bm{x},\bm{x}^{(b)})]^N_{b=1}$.
\begin{restatable}[KKR]{proposition}{KKR}\label{prop:KKR}
Consider a discrete-time dataset of Assumption \ref{ass:dataT}, $\Data^{\scale{\dt}}_N\tsgn{=}\{\bm{x}_{\trajH}^\datapt,{y}_{\trajH}^\datapt\}^N_{i=1}$, and let ${\bm{y}}^{\top}_{\trajH}{=}[{y}_{\trajH}^{(1)\top} {\cdots} {y}_{\trajH}^{(N)\top}]$ with ${\otimes}$ the Kronecker product. 
Then, 
\begin{align}\label{eq:LITsolv}
    \bm{\alpha}_j=k_{\bm{X}\naught \bm{X}\naught}^{-1}k_{\bm{X}_\trajH \bm{X}_\trajH}^{\deig_j} \left(\bm{I}_N\otimes \bm{\deig}_j^{* \top}\right) \underline{\bm{\beta}}, && \underline{\bm{\beta}} {=} (\bm{K}_{\bm{X}_{\trajH}\bm{X}_{\trajH}}{+}\gamma\bm{I}_{H\tsgn{+}1} \tsgn{\otimes} \bm{I}_N)^{-1}{\bm{y}}_{\trajH}
\end{align}
defines a unique time-sampled solution to \eqref{eq:LSreg} in terms of eigenfunctions $\hat{\bm{\phi}}(\bm{x}\naught) =  [{k}^{j}_{\bm{x}\naught\bm{X}\naught}
{\bm{\alpha}_j}
]
^{\bar{D}}_{j=1}$,
{determining} an LTI predictor\footnote{For discrete-time predictors, we omit the time-step specification and denote the next state with ``$(\cdot)^+$''.} with $\bm{\Lambda}=\operatorname{diag}([\mu_1 \cdots \mu_D])$,
\begin{subequations}\label{eq:DDlinearPredictor}
    \begin{align}
            {\bm{z}}^+ &= \bm{\Lambda} \bm{z},  ~\bm{z}\naught=\hat{\bm{\phi}}(\bm{x}\naught), \label{eq:ddLTI}\\
            {\hat{y}}&= \bm{1}^{\top}\!\bm{z} \label{eq:ddrecon}.
    \end{align}
\end{subequations} 
\end{restatable}
Notice how in \eqref{eq:LITsolv}, we re-scale the trajectory domain to that of the state-space. This enables us to write the forecast of \eqref{eq:DDlinearPredictor}, with a slight abuse of notation, using an extended observability matrix \cite{Qin2006AnIdentification}
\begin{equation}\label{eq:Rollout}
\hat{y}_{\trajH}=\bm{\mathrm{\Gamma}}\hat{\bm{\phi}}(\bm{x}\naught), \qquad \qquad \bm{\mathrm{\Gamma}}:=\left[\bm{1}^{\top}~\bm{1}^{\top} \bm{\Lambda}~\cdots 
~\bm{1}^{\top}\bm{\Lambda}^{H}  \right]^{\top}
.
\end{equation}

The confinement to a non-recurrent domain plays a crucial role in making the base kernel RKHSs
isometric to ``eigen-RKHSs'' $\RKHS^{k^{\scale{j}}} \cong\RKHS^{k^{\scale{\mu_j}}}$ via \textit{invariance transforms}, guaranteeing a feasible return from the time-series domain $\Set{X}_\trajT$ to the state-space domain $\Set{X}\naught \subseteq\Set{X}$ for evaluating the model over initial conditions.
\begin{remark}
    The salient feature of our proposed KKR framework compared to existing methods is the fact that Koopman-invariance \ref{eq:KIconst} over data samples is independent from the outcome of an optimization algorithm, e.g. minimizing the forecasting risk to compute $\underline{\bm{\beta}}$ in \eqref{eq:LITsolv}. Thus, we are able to directly optimize for a downstream task (forecasting) \ref{eq:lossRecon} given a suitably rich set of eigenvalues.
\end{remark}

\subsection{Selecting Eigenvalues}
Until now, we have used the sufficient cardinality $\bar{D} \in \Set{N}$ of an eigenvalue set that encloses \cite{Dunford1943SpectralProjections} or is the true spectrum.
However, we have provided no insight regarding the selection of $\bar{D}$ spectral components or how they can be estimated. Here, we go beyond the learning-independent and non-constructive existence result of \cite{Korda2020OptimalControl} and provide a consistency guarantee and relate it to sampling eigenvalues without the knowledge of the true spectrum.
\begin{restatable}{proposition}{KKconsistency}\label{prop:KKconv}
Consider the oracle Koopman kernel $\bm{K}(\bm{x}_{\trajH},{\bm{x}_{\trajH}}^\prime)$ and a dense set $\{\deig_j\}^{\infty}_{j=1}$ in $\overline{\Set{B}_1(\bm{0})}$. Then,
$\|\bm{K}(\bm{x}_{\trajH},{\bm{x}_{\trajH}}^\prime)-\textstyle{\sum^{{D}}_{j=1}} \bm{K}^{\deig_j}(\bm{x}_{\trajH},{\bm{x}_{\trajH}}^\prime)\|_{\mathcal{B}(\Set{Y}_{\trajH})} \rightarrow 0
$, $\forall~\bm{x}_{\trajH},{\bm{x}_{\trajH}}^\prime \in \Set{X}_{\trajH}$ as $D \rightarrow \infty$.
\end{restatable}
    As shown in Proposition \ref{prop:KKconv}, even if we do not know the \textit{oracle} kernel, we can arbitrarily approximate it by sampling from a dense set supported on the closed complex unit disk $\overline{\Set{B}_1(\bm{0})}$ \cite[Theorem 3.0.2]{Kuster2015TheSystems} with the error vanishing in the limit $D \!\rightarrow \! \infty$. There is no loss of generality when considering the unit disk as any finite radius disk can be scaled in the interval $[0,T]$. Furthermore, approximation of the oracle kernel by sampling a distribution over $\overline{\Set{B}_1(\bm{0})}$ leads to an almost sure $\mathcal{O}(\nicefrac{1}{\sqrt{D}})$ convergence rate.
    It is conceivable that faster rates can be obtained in practice by including prior knowledge to shape the spectral distribution, e.g. using well-known concepts such as leverage-scores or subspace  orthogonality \cite{Bach2017OnExpansions,Li2021TowardsFeatures}. Based on spectral priors one can include a more biased sampling technique by precomputing components of the operator spectrum, e.g. computing Fourier averages \cite{Mauroy2012OnDynamics}, to determine the phases $\omega_j$ of complex-conjugate pairs $\deig_{j, \pm} = |\deig_j|\operatorname{e}^{\pm \mathrm{i} \omega_j}$ and sample the modulus from another physics-informed distribution. 
    However, rigorous considerations of optimized and efficient sampling are beyond the scope of this paper and rather a topic of future work.
\subsection{Numerical Algorithm and Time-Complexity}
   \begin{wrapfigure}{L}{0.55\textwidth}
\vspace{-2\intextsep}
\begin{minipage}{0.55\textwidth}
  \begin{algorithm}[H]
    \caption{Regression and LTI Forecasts using KKR}
    \label{alg:KKR}
    \begin{algorithmic}
    \State Data $\Data\tsgn{=}\{\bm{x}_{\trajH}^\datapt,{y}_{\trajH}^\datapt\}^N_{i=1}$, Eigenvalues $\{\deig_j\}^{D}_{j=1}$
        \Function{Regress}{$\Data$, $\{\deig_j\}^{D}_{j=1}$}
            \State form Gramians $k_{\bm{X}\naught \bm{X}\naught},\{k_{\bm{X}_\trajH \bm{X}_\trajH}^{\deig_j}\}^D_{j=1},\bm{K}_{\bm{X}_\trajH \bm{X}_\trajH}$
            \State fit mode operator $\hat{{M}}(\cdot)\!:\Set{X}_\trajH\mapsto\Set{Y}_\trajH$ (\ref{eq:LITsolv}, right)
            \State recover eigenfunctions $\hat{\bm{\phi}}(\cdot)\!: \Set{X}_0\mapsto\Set{Z}\naught$ (\ref{eq:LITsolv}, left)
            \State construct $\bm{\mathrm{\Gamma}}\!: \Set{Z}\naught \mapsto \Set{Y}_\trajH$ (\ref{eq:Rollout}, right)
            \State \textbf{return} \textit{LTI predictor} $\bm{\mathrm{\Gamma}}\hat{\bm{\phi}}(\cdot): \Set{X}_0\mapsto\Set{Y}_\trajH$
        \EndFunction
        \Function{Forecast}{$\bm{x}_0$}
            \State ``lift'' $\bm{z}_0=\hat{\bm{\phi}}(\bm{x}_0)$ 
            \State rollout $\hat{y}_H=\bm{\mathrm{\Gamma}}\bm{z}_0$
            \State \textbf{return} trajectory $\hat{y}_H$
        \EndFunction
    \end{algorithmic}
  \end{algorithm}
\end{minipage}
\vspace{-0.5cm}
\end{wrapfigure}
For a better overview, the pseudocode for regression and forecasting of our method are shown in Algorithm \ref{alg:KKR}. We also put the time-complexity of our algorithm into perspective w.r.t. Koopman operator regression of PCR/RRR \cite{Kostic2022LearningSpaces} and ridge regression using state-of-the-art signature kernels \cite{Salvi2021ThePDE} (RR-Sig-PDE) in Table \ref{tab:complexity}. The training complexity of our KKR is comparable to that of RR-Sig-PDE regression and generally better than that of PCR/RRR. Given that accurate LTI forecasts require higher-rank predictors, the seemingly mild quadratic dependence makes $D^2 > NH$ and leads to a more costly matrix inversion. Furthermore, our LTI predictor also has a slightly better forecast complexity due to not depending on trajectory length. Obviously, due to a mere matrix multiplication after an initial nonlinear map, LTI predictors have a significantly lower evaluation complexity than the nonlinear predictor of Sig-PDE's. Due to requiring updated observation sequences as inputs, Sig-PDE kernels introduce a raw evaluation complexity that is also quadratic in sequence length.
    \begin{SCtable}[][h!]
    \vspace{-0.5\intextsep}
    \noindent\begin{minipage}[t]{1.55\linewidth}
      \centering
      \resizebox{\columnwidth}{!}{%
      \begin{tabular}{c|c|c}
        {Method} & Training & $H$-step forecast \\
        \hline
        KKR (ours) & $\mathcal{O}(N^3H^3\!+\!DN^2H^2d)$  
        &  $\mathcal{O}(DH\!+\!DNd)$  \\
        PCR/RRR     &  $\mathcal{O}(D^2N^2H^2\!+\!N^2H^2d)$  & $\mathcal{O}(DH\!+\!DNHd)$   \\
        RR-Sig-PDE   & $\mathcal{O}(N^3H^3\!+\!N^2H^2l^2d)$     & $\mathcal{O}(NH^2l^2d)$ \\
      \end{tabular}
      }
    \end{minipage}
    \noindent\begin{minipage}[t]{.45\linewidth}
      \centering
       \resizebox{1.05\columnwidth}{!}{%
      \begin{tabular}{c|c}
        $N$ & \# trajectories  \\
        $H$   &  trajectory length   \\
        $D$   & predictor rank  \\
        $l$& \# time-delays \\
        $d$ & $\mathrm{dim}$(input data) \\
      \end{tabular}
      }  
    \end{minipage}%
    \caption{Time complexities.}
    \label{tab:complexity}
    \vspace{-\intextsep}
\end{SCtable}

\section{Learning Guarantees}\label{section:lernGaran}
With a completely defined KKR estimator, we assess its essential learning-theoretic properties, i.e., the behavior of the learned functions w.r.t.\ to the ground truth with increasing dataset size.
\subsection{Consistency}\label{subsec:consistency}
 Although well-established in most function approximation settings \cite{Caponnetto2007OptimalAlgorithm,Carmeli2006VectorTheorem,Carmeli2010VectorUniversality}, the setting of Koopman-based LTI predictor learning for nonlinear systems is void of consistency guarantees.
  Here we use a definition of universal consistency from \cite{Caponnetto2008UniversalKernelsb} that describes the uniform convergence of the learned function to the target function as the sample size goes to infinity for any compact input space $\Set{X}$ and every target function $q\! \in \! C(\Set{X})$.
The existing convergence results for Koopman-based LTI predictors \cite{Korda2018OnOperator} are in the sense of strong operator topology --- allowing the existence of empirical eigenvalues that are not guaranteed to be close to true ones even with increasing data \cite{Rosenfeld2022}. This lack of spectral convergence has a cascaded effect in Koopman operator regression as, in turn, the convergence of eigenfunctions and mode coefficients is not guaranteed. Here, the convergence of modes is replaced by the convergence of eigenfunctions, and convergence of spectra is replaced by the convergence of \eqref{eq:Rollout} to the mode decomposition operator $\hat{M}\equiv\bm{\mathrm{\Gamma}}\hat{\bm{\phi}} \rightarrow M\!\equiv\!\bm{\mathrm{\Gamma}}{\bm{\phi}}$  with the estimate denoted by $\hat{(\cdot)}$.
\begin{restatable}[Universal consistency]{theorem}{universalConsistency}\label{theorem:universalConsistency}
    Consider a universal kernel $\bm{K}$ \eqref{eq:KEIGKernTrajDT2} and a data distribution supported on $\Set{X}_\trajH \tsgn{\times} \Set{Y}_\trajH$. Then,
    as $N \rightarrow \infty$,  $\|M
\tsgn{-}\hat{M}\|_{\Set{Y}_\trajH} \rightarrow 0$ and $\|\phi_{\deig_j}\tsgn{-}\hat{\phi}_{\deig_j}\|_{\Set{Y}_\trajH} \rightarrow 0, \forall j{=} {1},{\dots},{{D}}$.
\end{restatable}
{
\subsection{Generalization Gap: Uniform Bounds}
Due to formulating the LTI predictor learning problem as a function regression problem in an RKHS, we can utilize well-established concepts from statistical learning to provide bounds on the generalization capabilities of KKR. Given a dataset of trajectories, the following \textit{empirical risk }is minimized
$$\hat{\mathcal{R}}_N(\hat{M}) := \textstyle\frac{1}{N} \sum_{i \in [N]} \| y^\datapt_\trajH-\hat{M}(\bm{x}_\trajH^\datapt)\|_{\Set{Y}_\trajH}^2$$
which is ``in-sample'' mean square error (MSE) w.r.t. a trajectory-data generating distribution ${\rho_{\mathcal{D}}}$ of i.i.d. initial conditions. The \textit{true risk}/generalization error of an estimator is the ``out-of-sample'' MSE of the model on the entire domain and denoted as $\mathcal{R}(\cdot)$. Those quantities are, in essence, the model's performance on test and training data, respectively. Allowing for statements on the test performance with an increasing amount of data by means of training performance is a desirable feature in data-driven learning. Hence, we analyze our model in terms of the \emph{generalization gap} 
\begin{equation}\label{eq:generalizationGap}
|\mathcal{R}(\hat{M})-\hat{\mathcal{R}}_N(\hat{M})|=\left|\mathbb{E}_{(\bm{x}_\trajH,
{y}_\trajH) \sim \rho_{\mathcal{D}}}[{\| y_\trajH\tsgn{-}\hat{M}(\bm{x}_\trajH)\|_{\Set{Y}_\trajH}^2}]-\textstyle\frac{1}{N}\sum_{i=1}^{N}\| y^\datapt_\trajH\tsgn{-}\hat{M}(\bm{x}_\trajH^\datapt)\|_{\Set{Y}_\trajH}^2\right|.
\end{equation}
To ensure a well-specified problem, we require models in the hypothesis to admit a bounded norm.
\begin{restatable}[Bounded RKHS Norm]{assumption}{RKHSnormBound}\label{ass:RKHSnorm}
		The unknown function $M$ has a bounded norm in the RKHS $\dtRKHS$ attached to the Koopman kernel $\bm{K}(\cdot, \cdot)$, i.e., $\|M\|_{\dtRKHS} \leq B$ for some $B \in \mathbb{R}_{+}$.
	\end{restatable}
The above smoothness assumption is mild, e.g., satisfied by band-limited continuous trajectories \cite{Kanagawa2018GaussianEquivalences} and computable from data \cite{Scharnhorst2021RobustApproach,Csaji2022NonparametricFunctions}. In stark contrast, well-specified Koopman operator regression \cite{Kostic2022LearningSpaces} requires the operator to map the RKHS onto itself, which is a very strong assumption \cite{Das2020,Valva2023ConsistentCompactification}.

To derive the main result of this section, we utilize the framework of Rademacher random variables for measuring complexity of our model's hypothesis space, a concept generally explored in~\cite{Bartlett2002RademacherResultsc} and more particularly for classes of operator-valued kernels in~\cite{Huusari2021EntangledSeparability}. Conveniently, the derivation is, in terms of the RKHS $\dtRKHS$, similar to standard methods on RKHS-based complexity bounds~\cite{Bartlett2002RademacherResultsc}. We use well-known results based on concentration inequalities to provide high probability bounds on a model's generalization gap in terms of those complexities. Finally, we upper bound any constant with quantities specified in our assumptions and can state the following result.
\begin{restatable}[Generalization Gap of KKR]{theorem}{GeneralizationGapOfKKR}\label{theorem:generalizationRiskKKR}
	Let $\mathbb{D}^{\scale{\dt}}_N=\{\bm{x}_{\trajH}^\datapt, {y}_{\trajH}^\datapt\}_{i=1}^{N}$ be a dataset as in Assumption~\ref{ass:dataT} consistent with a 
	{Lipschitz system} on a 
	Then the generalization gap~\eqref{eq:generalizationGap} of a model $\hat{M}$ from Proposition~\ref{prop:KKR} under Assumption~\ref{ass:RKHSnorm} is, with probability $1-\delta$, upper bounded by
		\vspace{-1ex}
		\begin{equation}\label{theorem:generalizationRiskKKR:asymptotic} 
			\hspace{-1.375truecm}
			|\mathcal{R}(\hat{M})-\hat{\mathcal{R}}_N(\hat{M})|\leq
			4R B \sqrt{\frac{\kappa H^2}{N}}+\sqrt{\frac{8\log\frac{2}{\delta}}{N}}\in\mathcal{O}\left(\frac{H}{\sqrt{N}}\right),%
		\end{equation}
	where $R$ is an upper bound on the loss in the domain, and $\kappa$ the supremum of the base kernel. 
\end{restatable}

We observe an overall dependence of order {$\mathcal{O}(\nicefrac{1}{\sqrt{N}})$} w.r.t. data points, resembling the regular Monte Carlo rate to be expected when working with Rademacher complexities. 
Remarkably, an increase in the order of the predictor $D$ cannot widen the generalization gap but will eventually decrease the empirical risk due to the consistency of eigenspaces (Proposition \ref{prop:KKconv}).
Combined, our findings are a substantial improvement, both quantitatively and in terms of interpretability, over existing risk bounds on forecasting error~\cite[Theorem 1]{Kostic2022LearningSpaces}. Additionally, our intuitive non-recurrence requirement is easily verifiable from data. In contrast, the Koopman operator regression in RKHS comes with various strong assumptions \cite{Valva2023ConsistentCompactification} that require commonly unavailable expert knowledge.
Also, the generalization of existing Koopman-based statistical learning approaches depends on rank while ours is rank-independent. The significant implications of our results are demonstrated in the following.

\section{Numerical Experiments}\label{sec:numExp}
In our experiments\footnote{Additional details on the numerical experiments can be found in the supplementary material.}, we report the squared error of the forecast vector for the length of data trajectories averaged over multiple repetitions with corresponding min-max intervals. We validate our theoretical guarantees and compare to state-of-the-art operator and time-series approaches in RKHS.
For fairness, the same kernel and hyperparameters are chosen for our KKR, PCR (EDMD), RRR \cite{Kostic2022LearningSpaces} and regression with signature kernels (Sig-PDE) \cite{Salvi2021ThePDE}. Note, PCR and RRR are provided with the same trajectory data split into one-step data pairs while the time and observation time-delays are fed as data to the Sig-PDE regressor due to its recurrent structure. Along with code for reproduction of our experiments, we provide a {\ttfamily JAX} \cite{jax2018github} reliant Python module implementing a {\ttfamily sklearn} \cite{scikit-learn} compliant KKR estimator at {\urlstyle{rm}\url{https://github.com/TUM-ITR/koopcore}}.

{\bf Bi-stable system~} Consider an ODE $\dot{x} = ax+bx^3$ that arises in modeling of nonlinear friction. The parameters are $a=4$, $b=-4$, making for a bi-stable system at fixed points $\pm 1$. 
The numerical results are depicted in Figure~\ref{fig:Bernoulli3SamplesandH}. Sample trajectories both on training and testing data indicate the utility of the forecast risk minimization of KKR. While EDMD correctly captures the initial trend of most trajectories it fails to match the accuracy of Sig-PDE or our KKR predictors that utilize time-series structure. Furthermore, the behavior of KKR's generalization gap for an increasing time horizon $T=H\dt, \dt = {1}/{14}s$ closely matches our theoretical analysis.
\pgfplotsset{ticklabel style = {font=\scriptsize\sffamily},
	every axis label/.append style={font=\small\sffamily, yshift=6pt},
	legend style = {font=\scriptsize\sffamily},title style={yshift=-7pt, font = \small\sffamily} }
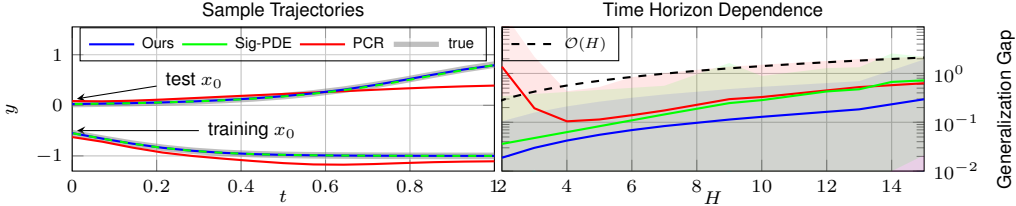
\begin{figure}
\centering
\begin{tikzpicture}
    \renewcommand{\pathtoresults}[0]{plots/data/bernoulli3/sampleTraj/}
        \begin{axis}[
                name=plot1,
                height = 3.5cm,
                width=0.515\textwidth,
                xmin=0, xmax=1,
                xlabel={\scriptsize $t$},
                xmajorgrids,
                ymin=-1.3, ymax=1.55,
                ylabel={\scriptsize $y$},
                yticklabel style={xshift=0pt},
                ylabel style={yshift=-10pt},
                ylabel near ticks, yticklabel pos=left,
                ymajorgrids,
                legend columns=4,
                legend style={
                nodes={scale=0.8, transform shape},
                column sep=1pt, text opacity=1,draw opacity=1,fill opacity=0.2,anchor=north west, xshift=-2pt, yshift=1.5pt},
                legend pos=north west,
                title={\scriptsize Sample Trajectories},
                   ylabel style={xshift=-10pt},
                    xlabel style={yshift=5pt},
            ]
            \addlegendimage{blue, thick}
            \addlegendentry{Ours};
            \addlegendimage{green, thick}
            \addlegendentry{Sig-PDE};
            \addlegendimage{red, thick}
            \addlegendentry{PCR};
            \addlegendimage{gray,opacity=0.5,line width=2pt}
            \addlegendentry{true};

            \node[anchor=west] (test) at (axis cs: 0.2,0.5) {\scriptsize\sffamily test ${x}_0$};
            \node[anchor=west] (train) at (axis cs: 0.3,-0.5) {\scriptsize\sffamily training ${x}_0$};
           \node[anchor=west] (testdest) at (axis cs:-0.033,0.1)
           {};
             \node[anchor=west] (traindest) at (axis cs:-0.033,-0.5){};
       \draw[-stealth](test)--(testdest);
       \draw[-stealth](train)--(traindest);
            \addplot[color=gray,opacity=0.5,line width=2.5pt] table[x=T,y=X]{\pathtoresults test_trajectory_r2_i44_10_data.csv};
            \addplot[color=red, thick] table[x=T,y=X]{\pathtoresults test_trajectory_r2_i44_10_theirs.csv};
            \addplot[color=blue, thick] table[x=T,y=X]{\pathtoresults test_trajectory_r2_i44_100_ours_isometric_.csv};
            \addplot[color=gray,opacity=0.5,line width=2.5pt] table[x=T,y=X]{\pathtoresults train_trajectory_r2_i36_data.csv};
            \addplot[color=red, thick] table[x=T,y=X]{\pathtoresults train_trajectory_r2_i36_10_theirs.csv};
            \addplot[color=blue, thick] table[x=T,y=X]{\pathtoresults train_trajectory_r2_i36_100_ours_isometric_.csv};
            \addplot[color=green, thick, dashed] table[x=T,y=X]{\pathtoresults train_trajectory_r2_i36_SK.csv};
            \addplot[color=green,thick, dashed] table[x=T,y=X]{\pathtoresults test_trajectory_r2_i44_SK.csv};
        \end{axis}
        \renewcommand{\pathtoresults}[0]{plots/data/bernoulli3/overH_mm/excess_risk_over_H_by_D}
        \begin{semilogyaxis}[
                name=plot2,
                at=(plot1.east),
                xshift=0.1cm,
                anchor=west,
                height = 3.5cm,
                width=0.515\textwidth,
                xmin=2, xmax=15,
                xlabel={\scriptsize $H$},
                xmajorgrids,
                ylabel={\scriptsize Generalization Gap},
                yticklabel style={xshift=0pt},
                ylabel style={yshift=-10pt},
                ylabel near ticks, 
                ymin=1e-2, ymax=9e0,
                yticklabel pos=right,
                ymajorgrids,
                legend columns=3,
                legend style={
                nodes={scale=0.8, transform shape},
                column sep=1pt, text opacity=1,draw opacity=1,fill opacity=0.2,
                at={(0.0,1.0)},
                anchor=north west},
                title={\scriptsize Time Horizon Dependence},
                   ylabel style={xshift=-10pt},
                   xlabel style={yshift=5pt},
            ]
            \addlegendimage{black, dashed, thick}
            \addlegendentry{$\mathcal{O}(\textstyle H)$};
            \addplot[
                domain = 1:15,
                samples = 15,
                smooth,
                thick,
                black,
                dashed,
            ] {x/sqrt(50)};
            \addriskplot{100}{ours_isometric_}{\pathtoresults}{blue}{blue}{thick}{}
            \addriskplot{10}{theirs}{\pathtoresults}{red}{red}{thick}{}
            \addriskplot{15}{SK}{\pathtoresults}{green}{green}{thick}{}
        \end{semilogyaxis}
\end{tikzpicture}
\vspace{-0.33cm}
\caption{Forecasting performance (48 i.i.d. runs) for the bi-stable system for $H\tsgn{=}14$ and $N\tsgn{=}50$ for respectively optimal $D_{\text{KKR}}\tsgn{=}100, D_{\text{PCR}}\tsgn{=}10$ and 15 delays for  Sig-PDEs. \textbf{Left:} Exemplary trajectories showing the advantage of learning with time-series kernels. \textbf{Right:} The generalization gap with an increasing forecast horizon, demonstrating generalization advantages of KKR.}
\label{fig:Bernoulli3SamplesandH}
\vspace{-0.2cm}
\end{figure}

{\bf Van der Pol oscillator~} 
Consider an ODE $\ddot{x}=\dot{x}(2-10x^2)-0.8x$ describing a dissipative system whose nonlinear damping induces a stable limit cycle --- a phenomenon present in various dynamics.
\begin{wraptable}[]{r}{0.6\textwidth} 
\vspace{-0.9\intextsep}
    \caption{Average risk  (20 runs) {\small${[\times 10^{-2}]}$} for Van der Pol for various \textit{spectral sampling} and \textit{lengthscales}, $N\tsgn{=}200$, $H\tsgn{=}14$.}
       \label{tab:sampling}
    \vspace{-0.33\intextsep}
 \resizebox{0.6\columnwidth}{!}{%
 \begin{tabular}{c|cc|cc|cc}
        $\rho(\mu)$ &  \multicolumn{2}{c|}{{uniform}} &  \multicolumn{2}{c|}{{boundary-biased}} &  \multicolumn{2}{c}{{physics-informed}}\\
        \hline
         $D$  & 16 & 200 & 16 & 200 & 16 & 200\\
        \hline
       $\mathcal{R}_{\ell\tsgn{=}10^1}$  & 13.7 & \textbf{5.38} & 11.2 & \textbf{5.38} & \colorbox{blue!20}{5.60} & 5.58\\
      $\mathcal{R}_{\ell\tsgn{=}10^0}$   & 6.46 & \textbf{0.78}  & 4.10 & \textbf{0.78} & \colorbox{blue!20}{0.97} & 0.92\\
      $\mathcal{R}_{\ell\tsgn{=}10^{-1}}$   & 7.12 & \textbf{1.74} & 4.33 & \textbf{1.74} & \colorbox{blue!20}{1.83} & 1.80
    \end{tabular}   
    }
    \vspace{-.7\intextsep}
\end{wraptable}
In Figure \ref{fig:NandDo} two fundamental effects are validated: the generalization gap with increasing data and consistency with test risk that does not deteriorate for increasing eigenspace cardinality.
The performance of PCR/RRR is strongly tied to predictor rank while Sig-PDE's less so w.r.t. delay length.
\input{plots/risk_van_der_pol}
\input{plots/risk_karman_hp}
\!\!\!\!\!\!\!\!\!\!\!\!\!\!\!\!\!\!\!\!\!\!\!\!Crucially, our KKR approach does not require a careful choice of the eigenspace cardinality to perform for a specific amount of data. 
Although the eigenvalues that determine the eigenspaces are randomly chosen from a uniform distribution in the unit ball, KKR consistently outperforms PCR/RRR.
In Table \ref{tab:sampling} we show the spectral sampling and hyperparameter effects. We employ the following strategies: 
 \textit{uniform} - uniform distribution on the complex unit disk, \textit{boundary-biased} - a distribution on the complex unit disk skewed towards the unit circle, \textit{physics-informed} - eigenvalues of various vector field Jacobians. As expected, physics-informed performs well with lower rank compared to uninformed approaches. However, it is outperformed by unit-ball sampling approaches for higher rank due to a lack of coverage. Table \ref{tab:compute_time} includes CPU timings for completeness.
 \vspace{-0.5\intextsep}
 \begin{SCtable}[][h]
 \begin{tabular}{c|cccc}
        $\#\text{data}=N\!\times\! H\!=\!200\!\times\!14$ & KKR & PCR & 
         RRR & Sig-PDE\\
        \hline
         Training $[\mathrm{s}]/$Forecast $[\mathrm{ms}]$ &  \textbf{8.0}/ \textbf{54}  & 90/ 84  & 88/150 & 8.6/5900 
    \caption{Computation times for Van der Pol.}  
    \label{tab:compute_time}
    \end{tabular}
\end{SCtable}

{\bf Flow~past~a~cylinder~}
We consider high-dimensional data of velocity magnitudes in a Kármán vortex street under varying initial cylinder placement, as illustrated in Figure~\ref{fig:karman}. 
 \begin{wrapfigure}{r}{0.38\textwidth}
\vspace{-1.1\intextsep}
  \begin{center}
    \includegraphics[width=0.38\textwidth]{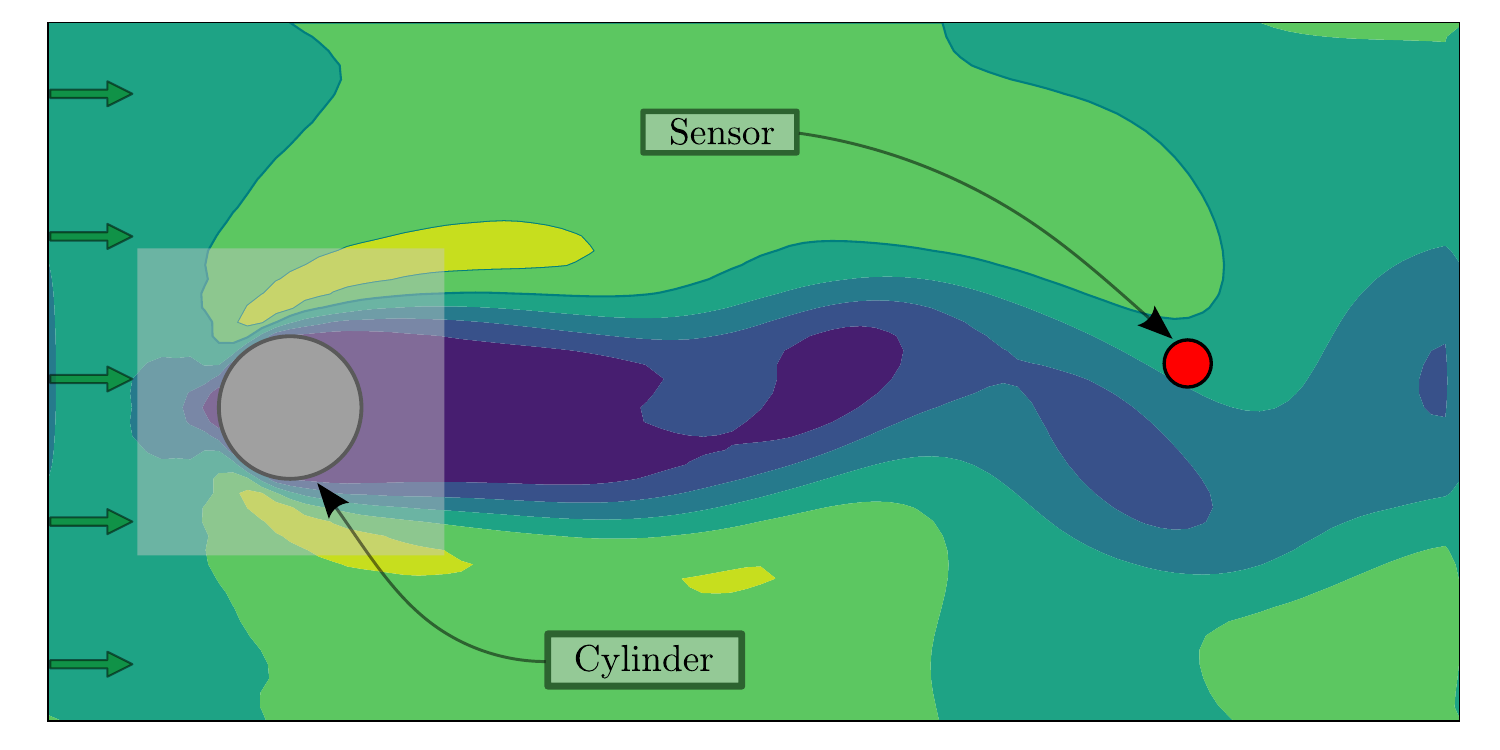}
  \end{center}
  \vspace{-0.6\intextsep}
  \caption{Flow illustration. Area of initial cylinder positions shaded.}
  \vspace{-0.9\intextsep}
  \label{fig:karman}
\end{wrapfigure}
The cylinder position is varied on a $7\tsgn{\times}7$ grid in a $50\tsgn{\times}100$-dimensional space and the flow is recorded over $H\tsgn{=}99$. The quantity of interest is a velocity magnitude sensor placed in the wake of the cylinder. In forecasting from an initial velocity field, KKR outperforms PCR by orders-of-magnitude as shown in Figure \ref{fig:accAndLen}. We omit Sig-PDE regression due to persistent divergence after $\approx$ 20 steps. The latter is hardly surprising, given that Sig-PDE models iterate one step predictions based only on the shapes of time-delays while LTI models directly output time-series from initial conditions.
\vspace{-0.5\intextsep}
\section{Conclusion}
We presented a novel statistical learning framework for learning LTI predictors using trajectories of a dynamical system. The method is rooted in the derivation of a novel RKHS over trajectories, which solely consists of universal functions that have LTI dynamics.
Equivalences with function regression in RKHS allow us to provide consistency guarantees not present in previous literature.
Another key contribution is a novel rank-independent generalization bound for i.i.d. sampled trajectories that directly describes forecasting performance.
The significant implications of the proposed approach are confirmed in experiments, leading to superior performance compared to Koopman operator and sequential data predictors in RKHS.
In this work, we confined our forecasts to a non-recurrent domain for a specific length of trajectory data, where the choice of spectra is arbitrary.
However, exploring more efficacious spectral sampling schemes is a natural next step for extending our results to asymptotic regimes that include, e.g., periodic and quasi-periodic behavior.
It has to be noted that vector-valued kernel methods have limited scalability with a growing number of training data and output dimensionality. Therefore, exploring solutions that improve scalability is an important topic for future work.
Furthermore, to enable the use of LTI predictors in safety-critical domains, the quantification of the forecasting error is essential. Hence, deriving uniform prediction error bounds for KKR is of great interest.\looseness=-1  
\newpage\subsection*{Acknowledgements}
The authors acknowledge the financial support of the EU Horizon 2020 research and innovation programme ``{SeaClear}'' (ID 871295) and the ERC Consolidator grant ``{CO-MAN}'' (ID 864686). Petar Bevanda also thanks Jan Br\"{u}digam for feedback, and Vladimir Kosti\'{c} and Pietro Novelli for useful discussions on operator regression.
\bibliographystyle{IEEEtran}
\bibliography{refs_cleaned}

\ifarxiv

\newpage
\appendix

\begin{center}
{\Large \bf Supplementary Material} 
\end{center}
\vspace{2em}

\renewcommand{\arraystretch}{1.33}
\begin{table}[!b]
\centering
\caption{Summary of used notation}
\label{tab:notation}
\begin{tabular}{c|c}
\hline
\toprule
Notation & Description   \\ 
\midrule 
$\Set{T}$ &  time interval $[0,T]$\\ \hline
$\Set{H}$ &  collection of points from discretizing the time interval $\Set{T}$ at times $\{t\naught, \cdots t_\trajH\}$\\ \hline
$\Set{X}$ &  compact state-space set \\ \hline
$\Set{X}_0$ &  compact set of initial conditions that form a non-recurrent domain\\ \hline
$\bm{x}_\trajT/\bm{x}_\trajH$ & a continuous/discrete time state trajectory \\ \hline
$\Set{X}_\trajT/\Set{X}_\trajH$ & space of continuous/discrete-time \textit{state} trajectories\\ \hline
${y}_\trajT/{y}_\trajH$ & a continuous/discrete time output trajectory \\ \hline
$\Set{Y}_\trajT/\Set{Y}_\trajH$ & space of continuous/discrete-time \textit{output} trajectories\\ \hline
$\ko^t$ & time-$t$ Koopman operator  \\ \hline
$M/\hat{M}$ & true/learned mode decomposition operator \\\hline
$K/\bm{K}/k$ &  operator/matrix/scalar-valued kernels \\\hline
$\eig/ \deig$ &  continuous/ discrete-time eigenvalue \\\hline
$K^{\eig_j}/\bm{K}^{\deig_j}/k^{j}$ &  operator/matrix/base kernel of the $j$-th Koopman eigenfunction \\\hline
$\RKHS^{k}$ & RKHS of a scalar base kernel $k$ \\ \hline
$\RKHS^{k^{\scale{\mu_j}}}$ & RKHS of a scalar kernel $k^{\mu_j}$ \\ \hline
$\RKHS^{\mu_j}$ & RKHS of matrix valued kernel $\bm{K}^{\deig_j}$ induced by scalar kernel $k^{\mu_j}$\\ \hline
$\RKHS^{\eig}$/$\RKHS^{\deig}$ & continuous/discrete-time Koopman eigenfunction RKHS $\eig/\deig \in \Set{C}$\\ \hline
$\RKHS^{}$/$\RKHS^{\dt}$ & continuous/discrete-time Koopman RKHS\\ \hline
$\mathcal{I}^{T}_{\eig}$/$\mathcal{I}^{H}_{\deig}$  & invariance transform for time/step length  $T$/$H$ and eigenvalue $\eig/\deig \in \Set{C}$\\ \hline
$\Data_{(\cdot)}$ & dataset for an estimator 
$(\cdot)$ \\\hline
$\Data_N$ & dataset of $N$ \textit{time-continuous} sample trajectories pairs $(\bm{x}_\trajT^\datapt,y_\trajT^\datapt)_{i\in[N]}$ \\\hline
$\Data^{\dt}_N$ & dataset of $N$ \textit{time-discrete} sample trajectories pairs $(\bm{x}_\trajH^\datapt,y_\trajH^\datapt)_{i\in[N]}$ \\ \hline
$\mathcal{B}(\cdot)$ & set of bounded operators over a domain \\\hline
$\overline{\Set{B}_r(\bm{0})}$ & closed ball of radius-$r$ in $\Set{C}$ \\\hline
$\bm{\Gamma}$ & extended observability matrix \\\hline
$\hat{\bm{\phi}}(\cdot)$ & vector-valued function of learned Koopman eigenfunctions\\\hline
$\mathcal{R}_N(\cdot)$ & true forecast risk/generalization error of an estimator \\\hline
$\hat{\mathcal{R}}_N(\cdot)$ & empirical forecast risk of an estimator based on $N$ data samples \\\hline
${R}_N(\cdot)$ & true Rademacher complexity of of a hypothesis class based on $N$ samples \\\hline
$\hat{R}_N(\cdot)$ & empirical Rademacher complexity of a hypothesis class based on $N$ samples \\\hline
$\mathcal{L}(\cdot)$ & loss function determining the metric for risk, e.g. squared error \\\hline
\bottomrule
\end{tabular}
\vspace{1em}
\end{table}

The supplementary material is organized as follows.
\begin{itemize}
    \item  Appendix~\ref{supl:koop} contains additional background on non-recurrence and spectral theory of Koopman operators. Additionally, it contains a notation table.  
    \item  Proofs of theoretical results are found in Appendix~\ref{supl:proof}.
\item Finally, Appendix~\ref{supl:exp} includes more details on the experimental section, as well as additional experiments.
\end{itemize}

\section{Non-recurrence and Koopman Operator Theory}\label{supl:koop}

\begin{remark}[Operator boundedness]\label{rmk:bounded}
    Consider a forward complete system on a compact set $\Set{X}$ and a continuous flow $\bm{F}^{t}$. It is well-known that a  time-$t$ Koopman operator $\ko^{t}$ is then a contraction semigroup on ${C}(\Set{X})$ \cite{Kreidler2018CompactSystems}. Due to forward completeness of the flow, we therefore obtain a Banach algebra ${C}(\Set{X})$ with a bounded semigroup $\{\ko^{t}\}_{t\geq0} \in \mathcal{B}({C}(\Set{X}))$.
\end{remark}
\begin{definition}[Non-recurrence]\label{rem:nonrecur}
    A non-recurrent domain is one where flow does not intersect itself.
\end{definition}
Non-recurrence is commonly ensured by a choice of the time interval $[0,T]$ so no periodicity is exhibited. Note that it does not mean the system's behavior is not allowed the be periodic, but our perception of it via data does. Effectively this prohibits the multi-valuedness of eigenfunctions -- allowing them to define an injective feature map.
    Thus, non-recurrence is a certain but general condition that bounds the time-horizon $T$ in which it is feasible to completely describe the nonlinear system's flow via an LTI predictor \eqref{eq:GenlinearPredictor}.
It makes for a less-restrictive and intuitive condition compared to existing RKHS approaches \cite{Kostic2022LearningSpaces,Kostic2023KoopmanEigenvalues} that rely on the self-adjointness and compactness of the actual Koopman operator which is rarely fulfilled and hard to verify without prior knowledge.
\begin{lemma}[Universality of Eigenfunctions]\label{lem:universal}
    Consider an quantity of interest $q \in C(\Set{X})$, a forward-complete system flow $\bm{F}^{t}(\cdot)$ on a non-recurrent domain $\Set{X}{\naught}$ (Definition \ref{rem:nonrecur}) of a compact set $\Set{X}$. Then, the output trajectory ${y}(t) = {q}(\bm{x}(t)), \forall t \in [0,T]$ is arbitrarily closely described by the eigenpairs $\{\eig_j,\phi_j\}_{j \in \Set{N}} {\subseteq} (\Set{C} \tsgn{\times} C(\Set{X}))$ of the Koopman operator semigroup $\{\ko^t\}^{T}_{t\tsgn{=}0}$\footnote{Note that, compared to ``Koopman Mode Decomposition'', we let the eigenfunctions absorb the spatial mode coefficients (possible w.l.o.g.) as they correspond to eigenfunctions and not eigenvalues \cite[Definition 9]{Budisic2012AppliedKoopmanism}.} so that $\forall \varepsilon > 0, \exists \bar{D} \in \Set{N}$
    \begin{equation}\label{modeDecom}
 |{q}(\bm{x}(t)) - \textstyle{\sum^{\bar{D}}_{j =1}}\operatorname{e}^{\eig_j t} \phi_j (\bm{x}\naught) | < \varepsilon, \forall t \in [0,T].
    \end{equation}
\end{lemma}
\begin{Proof}[Lemma \ref{lem:universal}]
With continuous eigenfunctions for continuous systems proved valid in \cite[Lemma 5.1]{Mezic2020SpectrumGeometry},\cite[Theorem 1]{Korda2020OptimalControl}, the space of continuous functions over a compact set is naturally the space of interest. On a non-recurrent domain, there exist uniquely defined non-trivial eigenfunctions and, by \cite[Theorem 3.0.2]{Kuster2015TheSystems}, the spectrum is rich -- with any eigenvalue in the closed complex unit disk legitimate \cite{Ikeda2022KoopmanSpaces}. Further, by \cite[Theorem 2]{Korda2020OptimalControl}, this richness is inherited by the Koopman eigenfunctions --- making them universal approximators of continuous functions.
\end{Proof}
\begin{remark}[Choosing the spectral distribution $\eig\sim \rho(\mu)$]\label{rmk:ChoiceOfMeasures}
    The choice of our measure of integration might seem arbitrary, and it indeed is. Since we, in general, do not assume knowledge of the spectrum of the Koopman-semigroup, we \textit{have to} make an approximation. To this end, an educated guess on where the (point-) spectrum might be located is helpful. As elaborated above, the Hille-Yosida-Theorem provides a convenient way to connect the practically attainable growth rates to bounds on the spectrum. Why would sampling spectral features in a set enclosing the spectrum be enough to obtain the spectral decomposition of the Koopman operator? Recalling that the spectral decomposition consists of projections to eigenspaces, we state a well-known result.
    The Riesz projection operator $P_\eig: \raum{C}\mapsto \{g\in\raum{C}: \operator{K}g=\eig g\}$ to an eigenspace of $\operator{K}$ can be represented by 
    \[P_{\eig} = \frac{1}{2\pi i}\normalint_{\gamma_{\eig}} \frac{\d{s}}{s - \operator{K}},\]
    where $\gamma_{\eig}$ is a Jordan curve enclosing $\eig$ and no other point in $\sigma(\operator{K})$ \cite{Dunford1943SpectralProjections}. 
    Obviously $\bigcup_{\eig\in\sigma(\operator{K})} \operatorname{range}(P_{\eig})=\raum{C}$, iterating on the fact that we can represent the operator $T$ by its spectral components.
    It becomes apparent that sampling from a set enclosing $\sigma(\eig)$ can be seen as sampling curves, eventually enclosing sufficient spectral components. And as stated, one can choose arbitrary measures on $\Complex$ as long as one ensures they enclose the spectrum. The preceding analysis sheds light on the connection of our approach to the Laplace-Stieltjes transform and the spectral pollution occurring in EDMD-type algorithms.
\end{remark}

\section{Proofs of Theoretical Results}\label{supl:proof}
{\bf Proof for Section~\ref{section:KKR} Koopman Kernel Regression~}
\begin{Proof}[Lemma~\ref{lemma:Invariants}]
 Due to the boundedness of finite-time trajectories of a forward complete system and a continuous $g\!\in C(\Set{X}_0)$ we have well-defined \textit{Haar integral invariants} \citep{Haasdonk2007InvariantLearning}
\begin{equation}\label{eq:efx0}
    \phi_{\eig}(\bm{x}_\trajT)  = \int^{T}_{\tau = 0} \operatorname{e}^{-\lambda (\tau-t)}\ko^{\tau}g(\bm{x}(0))d \tau \overset{}{=} \int^{T}_{0}\operatorname{e}^{-\lambda(\tau-t)} g(\bm{F}^\tau(\bm{x}_0))d \tau.
\end{equation}
Then, $\phi_{\lambda}\!:\Set{X}_0 \mapsto C(\Set{X}_0)$ \citep[p. 64]{Nachbin1976TheIntegral} is an invariant function for $\{\operatorname{e}^{-\eig \tau}\ko^\tau\}^{T}_{\tau=0}$ considering a normalized measure $d\tau(T)=1$ -- fulfilling the Koopman-invariance condition.
By simple algebraic manipulation we verify that $\phi_{\eig}$ indeed has LTI dynamics
\begin{align}
    \phi_{\eig}(\bm{x}_\trajT) &=  \int^{T}_{\tau=0}\operatorname{e}^{-\eig(\tau - t)} g(\bm{F}^{\tau}(\bm{x}\naught))d \tau\notag\\
    &=\operatorname{e}^{\eig t}\int^{T}_{\tau=0}\operatorname{e}^{-\eig\tau} g(\bm{F}^{\tau}(\bm{x}\naught))d \tau\notag\\
    &=\operatorname{e}^{\eig t}\phi_{\eig}(\bm{x}\naught).  \label{proofeq:algebaKI}
\end{align}
\end{Proof}
\begin{Proof}[Theorem~\ref{thm:eigKern}]\label{proof:theorem:eigKern}
    \begin{enumerate}[wide,labelindent = 0pt, labelwidth = !, label=(\roman*) ]
        \item              Due to the one-to-one relationship between kernel functions and RKHS we can examine $\RKHS^{\eig}$ by its kernel ${K}^{\eig}(\cdot, \cdot)$. We notice that due to the property that pointwise converging sequences of kernels are again kernels~\cite[Corollary 4.17]{IngoSteinwart2008SupportMachines}. Showing that ${K}^\eig$ is a kernel thus reduces to showing that the double integral exists.
 Now, since our continuity assumptions on the system ensure the convergence of the Haar-integrals~\citep[p. 64]{Nachbin1976TheIntegral}, we can conclude that any valid integration scheme~\cite[Theorem A.1.5]{Werner2018Funktionalanalysis} induces a uniformly converging sequence of kernels.
        \item         We will prove the statement by showing that the universality of the base kernel for continuous functions makes the Koopman eigenfunction RKHS $\RKHS^\lambda$ universal for continuous Koopman-invariant functions at eigenvalue $\lambda \in \Set{C}$.
It is clear that feature map of the kernel is $\{\operatorname{e}^{-\eig \tau}\ko^\tau\}^{T}_{\tau=0}$-invariant, and we only need to prove the completeness part. Let $\Set{X}\naught$ be a compact subset in $\Set{X}$, and $\epsilon>0$. Then, the non-recurrent domain defined by $\Set{X}_{{T}}=\cup_{t \in [0,T]} \bm{F}^t({\Set{X}_0})$ under the continuous map $(t, \bm{x}) \mapsto \bm{F}^t(\bm{x})$ is also a compact set. By using a universal RKHS $\RKHS^k$, we know there exists $f \in \RKHS^k$ so that 
$$\sup_{\bm{x} \in \Set{X}_T}\left|f(\bm{x}) - \phi_{\lambda}(\bm{x})\right| \leq \epsilon.$$ 
Consider now a $\{\operatorname{e}^{-\eig \tau}\ko^\tau\}^{T}_{\tau=0}$-invariant group-averaged map ${f}_\lambda(\bm{x})=\normalint^T_{\tau=0} \operatorname{e}^{-\lambda \tau}f\left(\bm{x}(\tau))\right) d \tau$ from the Koopman eigenfunction RKHS $\RKHS^\lambda$ induced by Lemma 1. Then due to
$$
\begin{aligned}
\sup_{\bm{x} \in \Set{X}\naught}\left|{f}_\lambda(\bm{x})- \phi_{\lambda}(\bm{x})\right| & =\sup_{\bm{x} \in \Set{X}\naught}\left|\int^{T}_{\tau=0}\left(\operatorname{e}^{-\lambda \tau} f\left(\bm{x}(\tau)\right)-\operatorname{e}^{-\lambda \tau}  \phi_{\lambda}\left(\bm{x}(\tau)\right)\right) d\tau\right| \\
\text{({\text{triangle inequality}})} \quad & {\leq}\sup_{\bm{x} \in \Set{X}\naught}\int^{T}_{\tau=0}\left|\left(\operatorname{e}^{-\lambda \tau} f\left(\bm{x}(\tau)\right)-\operatorname{e}^{-\lambda \tau}  \phi_{\lambda}\left(\bm{x}(\tau)\right)\right)\right| d\tau \\
 & \leq \int^{T}_{\tau=0}\sup_{\bm{x} \in \Set{X}\naught}\left|\left(\operatorname{e}^{-\lambda \tau} f\left(\bm{x}(\tau)\right)-\operatorname{e}^{-\lambda \tau}  \phi_{\lambda}\left(\bm{x}(\tau)\right)\right)\right| d\tau \\
 \text{({\text{Cauchy–Schwarz inequality}})} \quad  & \leq \int^{T}_{\tau=0}\left|\operatorname{e}^{-\lambda \tau}\right|\sup_{\bm{x} \in \Set{X}\naught}\left| f\left(\bm{x}(\tau)\right)- \phi_{\lambda}\left(\bm{x}(\tau)\right)\right| d\tau \\
& \leq 
  \sup _{\tau^\prime \in [0,T]}\left|\operatorname{e}^{-\lambda \tau^\prime}\right| \int^{T}_{\tau=0} \sup_{\bm{x} \in \Set{X}_T}\left|f\left(\bm{x}\right)- \phi_{\lambda}\left(\bm{x}\right)\right| d\tau \\
& =  \max\{1,\left|\operatorname{e}^{-\lambda T}\right|\} T \epsilon,
\end{aligned}
$$
we can approximate any Koopman eigenfunction $\phi_\lambda$ with a Koopman-invariant function ${f}_\lambda$ to arbitrary accuracy.
        \item  With the knowledge of an explicit LTI feature representation from Lemma 1, we show that $\RKHS^{\lambda}$ satisfies Koopman-invariance along sampled trajectories $\{\bm{x}^\datapt_{\trajT}\}_{i=1}^N$. For representing an open eigenfunction over an initial condition, we choose an RKHS $\RKHS^k$ of a universal kernel $k(\cdot,\cdot)\!:~\Set{X} \times \Set{X} \mapsto \Set{R}$. As a consequence of Mercer's theorem \cite{Mercer1909FunctionsEquations}, there exists a feature map $\bm{\xi}\!:\Set{R}^{d} \mapsto \RKHS^k$ for every kernel $k(\cdot,\cdot)$ such that 
\begin{equation}\label{eq:kernx0EF}
    k(\cdot,\cdot)= \langle \bm{\xi}(\cdot),\bm{\xi}(\cdot)\rangle_{\RKHS^k}.
\end{equation} 
Due to universality of $k(\cdot,\cdot)$ and continuity of eigenfunctions \cite{Mezic2020SpectrumGeometry}, there exists a parameter vector $\bm{\theta}$ so that
        \begin{align}
            g(\bm{x}_{\trajT}^\datapt(0)) & =\langle\bm{\theta}, \bm{\xi}(\bm{x}_{\trajT}^\datapt(0))\rangle_{\RKHS^k}, \quad \forall i=1, \ldots, N .
        \end{align}
To enforce Lemma 1 at data points we utilize an RKHS $\RKHS^{\eig}$
induced by $\mathcal{I}^{T}_{\eig}\!:\RKHS^k \rightarrow \RKHS^{\eig}$. Due to universality for arbitrary continuous Koopman eigenfunctions by $(ii)$, there exists a parameter vector $\bm{\alpha}$ so that
    \begin{align}
                    f_{\eig}(\bm{x}_{\trajT}^\datapt) & = \langle\bm{\alpha}, \mathcal{I}^T_{\eig}\bm{\xi}(\bm{x}_{\trajT}^\datapt(0))\rangle_{\RKHS^{\eig}}, \quad \forall i=1, \ldots, N .\label{eq:FeatureLoperator}
    \end{align}
From \eqref{eq:FeatureLoperator} we recognize a modified feature map $\bm{\psi}(\cdot) = \mathcal{I}^T_{\eig}\bm{\xi}(\cdot)$, representing the eigenfunction flow at $\bm{x}_{\trajT}^\datapt, i=1,\dots,N$, $\forall t \in [0,T]$
\begin{align}
    f_{\eig}(\bm{x}_\trajT) & = \langle\bm{\alpha}, \bm{\psi}(\bm{x}_{\trajT}^\datapt)\rangle_{\RKHS^{\eig}}, \quad \forall i=1, \ldots, N,
\end{align}
inducing a kernel 
\begin{equation}
    K^{{\eig}}(\cdot,\cdot) = \langle \bm{\psi}(\cdot),\bm{\psi}(\cdot)\rangle_{\RKHS^{\eig}}.
\end{equation}
By exploiting inner product properties, we recognize
\begin{equation}
    K^{{\eig}}(\cdot,\cdot) = \langle  \mathcal{I}^T_{\eig}\bm{\xi}(\cdot), {\mathcal{I}^T_{\eig}}\bm{\xi}(\cdot)\rangle_{\RKHS^\eig},
\end{equation}
leading to
    \begin{equation}
    K^{{\eig}}(\bm{x}_{\trajT},\bm{x}_{\trajT}^{\prime}) =  \mathcal{I}^T_{\eig} (\mathcal{I}^{T}_{\eig})^* \langle \bm{\xi}(\bm{x}_{\trajT}(0)),\bm{\xi}(\bm{x}_{\trajT}^{\prime}(0)\rangle_{\RKHS^k}  = \mathcal{I}^T_{\eig} k(\bm{x}_{\trajT}(0),\bm{x}_{\trajT}^{\prime}(0)) \mathcal{I}^{T^{\prime}}_{\eig^*}.
\end{equation}
Finally, by applying the operators to the kernel, we obtain the induced ``Koopman kernel''
\begin{equation}\label{keigEQ}
K^{{\eig}}(\bm{x}_{\trajT},\bm{x}_{\trajT}^{\prime}) =\int^T_{\tau=0} \int^T_{{\tau}^{\prime}=0} \frac{k\left(\bm{x}_{\trajT}^{}(\tau),\bm{x}_{\trajT}^{\prime}(\tau^{\prime}))\right)}{\operatorname{e}^{{\eig} (\tau-t)}\operatorname{e}^{\eig^*(\tau^{\prime}-t)}} d \tau d\tau^{\prime}.
\end{equation}
fulfilling Lemma 1 along sampled trajectories $\bm{x}_{\trajT}^\datapt,$ $i,\ldots,N$.    
    \end{enumerate}  
\end{Proof}
\begin{Proof}[Proposition~\ref{prop:koKern}]\label{proof:Prop1}
    \begin{enumerate}[wide,labelindent = 0pt, labelwidth = !, label=(\roman*)  ]
        \item We show that $\RKHS$ is an RKHS by showing it is associated with a kernel which is the limit of a pointwise converging sequence of kernels~\cite[Corrollary 4.17]{IngoSteinwart2008SupportMachines}. Since ${K}^\eig$ is a finite sum, it is bounded by virtue of its elements being bounded, which is due to Theorem~\ref{thm:eigKern},(i). 
        \item Universality of $\RKHS$ is guaranteed by using eigenspace universality \cite[Theorem 2]{Korda2020OptimalControl} and applying Theorem~\ref{thm:eigKern}~(ii) component-wise. Our goal is to represent a function in terms of an LTI predictor, the mode composition of the Koopman operator. Due to Lemma \ref{lem:universal}, we know the exact mode decomposition $\ModOp$ is countable so the contribution of neglected eigenspaces can be made arbitrarily small by choosing $\bar{D}$ large enough.
        \begin{align*} \
            \|{y}_\trajT - \hat{\ModOp}(\bm{x}_\trajT)\|_{\Set{Y}_\trajT}
            &=\|{\ModOp}(\bm{x}_\trajT) - \hat{\ModOp}(\bm{x}_\trajT)\|_{\Set{Y}_\trajT}\\
            &=\|\bm{1}^\top [\phi_{\eig_1}\cdots\phi_{\eig_{\bar{D}}}](\bm{x}_\trajT)-\bm{1}^\top [\hat{\phi}_{\eig_1}\cdots\hat{\phi}_{\eig_{{D}}}\cdots](\bm{x}_\trajT)\|_{\Set{Y}_\trajT}\\
            &=\|\phi_{\eig_{1}}-\hat{\phi}_{\eig_{1}}+\cdots+\phi_{\eig_{\bar{D}}}-\hat{\phi}_{\eig_{\bar{D}}}+\sum_{j=\bar{D}+1}^{\infty}\phi_{\eig_{j}}\|_{\Set{Y}_\trajT}\\
            &\leq \|\phi_{\eig_{1}}-\hat{\phi}_{\eig_{1}}\|_{\Set{Y}_\trajT}+\cdots+\|\phi_{\eig_{\bar{D}}}-\hat{\phi}_{\eig_{\bar{D}}}\|_{\Set{Y}_\trajT}+\delta\\
            &\overset{\text{Proposition~\ref{def:KEIGS}~(ii)}}{\leq}\epsilon_{1}+\cdots+{\epsilon_{{\bar{D}}}}+\delta
        \end{align*}
        Now choosing $\bar{D}$ such that $\delta<\epsilon$ and $\epsilon_i=\frac{\epsilon -\delta}{\bar{D}}$, yields the assertion.
    \end{enumerate}
\end{Proof}

\begin{Proof}[Corollary \ref{cor:trajK}]
\begin{enumerate}[wide,labelindent = 0pt, labelwidth = !, label=(\roman*) ]
    \item By considering the integral equation \eqref{eq:KEIGKern} at $H$ regular intervals $\dt$ so that $H = T/\dt$ with $\forall t \in \{t_k\}^{H}_{k=0}$ the integrals are replaced by sums. Due to considering normalized measures of $d\tau(T)$ and $d\tau^{\prime}(T)$ in \eqref{eq:KEIGKern}, each sum is normalized by the number of elements $(H+1)$, resulting in \eqref{eq:KEIGKernTrajDT}. All properties from Theorem \ref{thm:eigKern} transfer straightforwardly using the same arguments as in Proof \ref{proof:theorem:eigKern}. 
    \item The construction of the kernel matrix sum directly follows directly follows the direct Hilbert space sum
    \begin{equation}
            \tilde{\RKHS}^\dt = {\RKHS^{\deig_1} \oplus\cdots \oplus\RKHS^{\deig_{\bar{D}}}} ~~ \operatorname{so~that} ~~ \RKHS^\dt = \operatorname{range}(\mathcal{S}){:=} \{ f_{1}+\ldots+f_{\bar{D}}: f_{1} \tsgn{\in} \RKHS^{\deig_1},\ldots,f_{\bar{D}}\tsgn{\in}\RKHS^{\deig_{\bar{D}}} \}
    \end{equation}
    All properties straightforwardly transfer from Proposition \ref{prop:koKern} using the same arguments as in Proof \ref{proof:Prop1}.
\end{enumerate}
\end{Proof}

\begin{Proof}[Proposition \ref{prop:KKR}]

It is easily recognizable that the time-discretization of problem \eqref{eq:LSreg} reads 
 \begin{equation}\label{eq:RepThm}
    \operatornamewithlimits{min}_{\scriptstyle\underline{\bm{\beta}}^\top=[{\bm{\beta}}_1 \cdots {\bm{\beta}}_N]}\textstyle\sum_{i=1}^N\|{y}_{\trajH}^\datapt - \bm{K}({\bm{x}_{\trajH}^\datapt,\bm{X}_{\trajH}}){\bm{\beta}}_i\|_{\Set{Y}_\trajH}^2+{\gamma}{\bm{\beta}}_i^{\top}\bm{K}(\bm{x}_{\trajH}^\datapt,\bm{x}_{\trajH}^\datapt){\bm{\beta}}_i.
\end{equation}
with
     $\underline{\bm{\beta}}$ the unique solution to the system of linear equations
\vspace{-.15truecm}
\begin{equation}
\left(\bm{K}(\bm{X}_{\trajH},\bm{X}_{\trajH})+\gamma \bm{I}_{H+1} \otimes \bm{I}_N\right)\underbrace{[\bm{\beta}^{\top}_1, \ldots, \bm{\beta}^{\top}_N]^{\top}}_{\underline{\bm{\beta}}}=\underbrace{[{y}_{\trajH}^{(1)^\top}, \ldots, {y}_{\trajH}^{(N)^\top}]^{\top}}_{\bm{y}_\trajH},
\end{equation}
Due to being a particular case linear coregionalization models \cite{Lederer2021TheControl,Alvarez2011KernelsReview}, it follows that the approximations $\hat{\phi}_j(\cdot)$ of Koopman eigenfunctions satisfying Definition \ref{def:KEIGS} over trajectory samples $\{\bm{x}_{\trajH}^\datapt\}^{N}_{i=1}$ are uniquely defined by
    \begin{equation}\label{eq:trajKEIGS}
         \hat{\phi}_j(\bm{x}_\trajH) =  \sum_{i=1}^N \left(k^{{{\deig}_j}}\left(\bm{x}_\trajH,\bm{x}_{\trajH}^\datapt\right) \otimes \bm{\deig}_j^{*\top}\right)\bm{\beta}_i = k_{\bm{X}_\trajH \bm{X}_\trajH}^{\deig_j} \left(\bm{I}_N\otimes \bm{\deig}_j^{* \top}\right) \underline{\bm{\beta}}.
    \end{equation}
As a consequence of a non-recurrent domain, the time-discrete invariance transformation
is a bijection at time-instances of the trajectory. Therefore, a base kernel RKHS $\RKHS^{k^{\scale{j}}}$ is isometrically isomorphic to $\RKHS^{k^{\deig_j}}$ with isometry $\mathcal{I}^{H}_{\deig_j}$, it is guaranteed $\forall \bm{x}_\trajH \in \Set{D}_N^\dt \mid \bm{x}\naught\equiv\bm{x}_\trajH(0)$
\begin{subequations}
    \begin{align}
         \hat{\phi}_j(\bm{x}\naught) &= \hat{\phi}_j(\bm{x}_\trajH), \\
           k^j_{\bm{X}_0\bm{X}_0} \bm{\alpha}_j &=  k_{\bm{X}_\trajH \bm{X}_\trajH}^{\deig_j} \left(\bm{I}_N\otimes \bm{\deig}_j^{* \top}\right) \underline{\bm{\beta}}.
    \end{align}
\end{subequations}
Then via $\bm{\alpha}_j=k_{\bm{X}\naught \bm{X}\naught}^{-1}k_{\bm{X}_\trajH \bm{X}_\trajH}^{\deig_j} \left(\bm{I}_N\otimes \bm{\deig}_j^{* \top}\right) \underline{\bm{\beta}}$ eigenfunctions are uniquely determined as
    \begin{equation}
        \hat{\bm{\phi}}(\bm{x}\naught) =  \left[{k}^{j}_{\bm{x}\naught\bm{X}\naught}
        {\bm{\alpha}_j}
        \right]
        ^{\bar{D}}_{j=1},
    \end{equation}
    concluding the proof.

\end{Proof}
\begin{Proof}[Proposition~\ref{prop:KKconv}]
Due to \cite[Theorem 3.0.2]{Kuster2015TheSystems}, we consider, w.l.o.g., a dense set $\{\deig_j\}^{\infty}_{j=1}$ in $\overline{\Set{B}_1(\bm{0})}$ and a finite-rank kernel
$\tilde{\bm{K}}=\textstyle{\sum^{{D}}_{j=1}} \bm{K}^{\deig_j}(\bm{x}_{\trajH},{\bm{x}_{\trajH}}^\prime)$. As the ``oracle'' kernel ${\bm{K}}=\textstyle \normalint_{\deig \sim \rho({\overline{\mathbb{B}_1(\bm{0})}})}\bm{K}^{\deig}(\bm{x}_{\trajH},{\bm{x}_{\trajH}}^\prime) \d{\deig}$ is an operator norm limit of compact Riemann sums $\tilde{\bm{K}}$ on a Hilbert space $\Set{Y}_{\trajH}$, it is a compact operator. Thus, by \cite[ Theorem II (p. 374)]{Aronszajn1950TheoryKernels}, $\tilde{\bm{K}} \rightarrow {\bm{K}}$ uniformly as $D \rightarrow \infty$. 
\end{Proof}
\begin{Proof}[Theorem \ref{theorem:universalConsistency}]\label{proof:thm:univConst}
    Consider a universal Koopman kernel $\bm{K}$. Consider the base kernel is Mercer and recall the properties of the invariance transformation from the proof of Corollary \ref{cor:trajK}: the matrix-valued kernel $\bm{K}$ is trace-class as $\mathcal{I}^H_\deig \mathcal{I}^{H*}_\deig$ is a bounded self-adjoint operator \cite{Carmeli2006VectorTheorem} and the base kernel is Mercer 
    \cite{Mercer1909FunctionsEquations}. With Proposition \ref{prop:KKconv}, the universal consistency is immediate via \cite{Caponnetto2008UniversalKernelsb}. Thus, as $N \rightarrow \infty$, the mode decomposition is consistent $\|M-\hat{M}\|_{\Set{Y}_\trajH} \rightarrow 0$ and the same immediately follows for individual eigenfunctions as the universality of summand RKHSs is unaffected so $\|\phi_{\deig_j}-\hat{\phi}_{\deig_j}\|_{\Set{Y}_\trajH} \rightarrow 0$, $j=1,\dots,
    \overline{D}$.
\end{Proof}

{\bf Proofs for Section~\ref{section:lernGaran} Generalization Gap: Uniform Bounds~}
We use the seminal result of~\cite{Bartlett2002RademacherResultsc}, which we will restate here for completeness.
\begin{theorem}[Rademacher Generalization Risk Bound, \cite{Bartlett2002RademacherResultsc} -- Theorem 8, 11]\label{theorem:RademacherGeneralizationRiskBound}
    Consider a loss function $\loss: \mathcal{Y}\times\mathcal{A}\to[0, 1]$. Let $\mathcal{F}$ be a class of functions with signature $\mathcal{X}\to\mathcal{A}$ and let $\{X_i, Y_i\}^N_{i=1}$ be independently selected according to the probability measure P. then, for any integer $n$ and any $\delta\in \left(0, 1\right)$, with probability at least $1-\delta$ over samples of length $n$, every $f\in \mathcal{F}$ satisfies
    \begin{align*}
    \mathbb{E}[\loss (Y, f(X))]
    &\leq
    \hat{\Erw}^\scale{N}[{\loss}(Y, f(X))]+2L({\loss}\naught)R_\scale{N}(\mathcal{F})+\sqrt{\frac{8\log\frac{2}{\delta}}{N}}
    ,
    \end{align*}
where ${\loss}\naught(y, a) = {\loss}(y, a)-\tilde{\loss}(y, 0)$.
\end{theorem}
To apply it to our use-case, we need to quantify the Rademacher complexities of our hypothesis space for which we make the following assumption.
\RKHSnormBound*
An extension of classical results for operator-valued Rademacher complexities:
\begin{restatable}[Rademacher Complexities of the Koopman Kernel]{lemma}{RademacherComplexitiesOfKoopmanKernel}\label{lemma:RademacherComplexitiesKK} Consider the, Mercer, Koopman kernel $\bm{K}$ and $\dtRKHS$ its RKHS as defined Corollary~\ref{cor:trajK} and $T_{\bm{K}} g=\normalint_{\mathbb{X}_{\trajH}} \bm{K}(\cdot, \bm{{x}}_{\trajH}) g(\bm{{x}}_{\trajH})\d{\bm{\tilde{x}}_{\trajH}}$ the corresponding integral operator on $L^2(\Set{X}_{\trajH})$. Then under Assumption \ref{ass:RKHSnorm}, the Rademacher complexities of $\dtRKHS$ are upper bounded by  
\\\noindent\begin{tabular*}{\textwidth}{cc}
    {\rm Asymptotic:~}$R_{\scale{N}}(\RKHS^{\dt})\leq \frac{B}{\sqrt{N}}\sqrt{\operatorname{trace}\left({T}_{\bm{K}}\right)}$
    &
    {\rm Non-Asymptotic:~}$\hat{R}_\scale{N}(\RKHS^{\dt})\leq\frac{B}{N}\sqrt{\operatorname{trace}\left(T_{\bm{K}}^\scale{N}\right)}$,
\end{tabular*}
\end{restatable}
\begin{Proof}[Lemma~\ref{lemma:RademacherComplexitiesKK}]
We derive an upper bound on the Rademacher complexities of the Koopman kernel using a procedure similar to the one described in~\cite[Lemma 22] {Bartlett2002RademacherResultsc}. Let $X_i$ be random element of $(\Set{X}_\trajH,\rho_{\mathcal{D}})$ and $\sigma$ a vector of independent uniform random functions on $\{-1, 1\}$, then the $n$-th Rademacher complexity of $\mathcal{F}$ is defined as
    \begin{align*}
        R_\scale{N}(\raum{F}) 
        =
        \mathbb{E}_{\sigma, \rho_{\mathcal{D}}} \sup\limits_{f\in \mathcal{F}}\frac{1}{N}\sum\limits_{i=1}^{N}|\langle\sigma_i ,f(X_i)\rangle|
        \overset{\text{scalar}}{=}
        \mathbb{E}_{\sigma, \rho_{\mathcal{D}}} \sup\limits_{f\in \mathcal{F}}\frac{1}{N}\sum\limits_{i=1}^{N}\sigma_if(X_i)
        .
    \end{align*}
    The empirical case $\hat{R}_n$ is similar to the expectation of $\sigma$.
    Now consider the Rademacher complexities of the RKHS $\RKHS^\dt$ corresponding to the Koopman kernel for some fixed $D$, with respect to initial conditions $\bm{x}^\datapt_\trajH$ drawn from $(\Set{X}_\trajH, \rho_{\mathcal{D}})$.
    \begin{align*}
        R_\scale{N}(\RKHS^{\dt}_N)
        &=
        \Erw_{\sigma, \rho_{\mathcal{D}}}\sup\limits_{M\in\dtRKHS_N}\frac{1}{N}\sum\limits_{i=1}^{N}|\langle\sigma_i ,M(\bm{x}_{\trajH}^{(i)})\rangle|
        &%
        \\
        &\leq
        &
        \\
        R_\scale{N}(\RKHS^{\dt})
        &=
        \Erw_{\sigma, \rho_{\mathcal{D}}}\sup\limits_{M\in\dtRKHS}\frac{1}{N}\sum\limits_{i=1}^{N}|\langle\sigma_i ,M(\bm{x}_{\trajH}^{(i)})\rangle|
        &\textit{Pre-RKHS property}
        \\
        &\leq
        \Erw_{\sigma, \rho_{\mathcal{D}}}\sup\limits_{M\in\dtRKHS}\frac{1}{N}\sum\limits_{i=1}^{N}\|\sigma_i\|_2 \|M(\bm{x}_{\trajH}^{(i)})\|_2
        &\textit{H\"older's inequality}
        \\
        &=
        \Erw_{\rho_{\mathcal{D}}}\sup\limits_{M\in\dtRKHS}\frac{1}{N}\sum\limits_{i=1}^{N} \|M(\bm{x}_{\trajH}^{(i)})\|_2
        &\textit{property of Rademacher functions}
        \\
        &\leq
        \Erw_{\rho_{\mathcal{D}}}\sup\limits_{\|\underline{\bm{\beta}}\|\leq B}\frac{1}{N}\sum\limits_{i=1}^{N} \| \bm{K}(\bm{x}_{\trajH}^{(i)}, \cdot) \bm{\beta} \|_2
        &\textit{by construction}
        \\
        &\leq
        \Erw_{\rho_{\mathcal{D}}}\frac{1}{N}\sum\limits_{i=1}^{N}  B\|\bm{K}(\bm{x}_{\trajH}^{(i)}, \cdot)\|_2
        &\textit{operator norm}
        \\
        &=
        \Erw_{\rho_{\mathcal{D}}}\frac{B}{N}\sum\limits_{i=1}^{N}\sqrt{\bm{K}(\bm{x}_{\trajH}^{(i)}, \bm{x}_{\trajH}^{(i)})}
        &\textit{reproducing property}
    \end{align*}
    By applying concavity and the respective definition, it follows that 
    \begin{align*}
    R_\scale{N}(\RKHS^{\dt})
    &\leq \frac{B}{\sqrt{N}}\sqrt{\frac{1}{N}\sum\limits_{i=1}^{N}\Erw_{\rho_{\mathcal{D}}}\bm{K}(\bm{x}_{\trajH}^{(i)}, \bm{x}_{\trajH}^{(i)})}=\frac{B}{\sqrt{N}}\sqrt{\operatorname{trace}\left( T_{\bm{K}}\right)} 
    \\
    \text{and}&\\
    \hat{R}_\scale{N}(\RKHS^{\dt})
    &\leq\frac{B}{N}\sum\limits_{i=1}^{N}\sqrt{\bm{K}(\bm{x}_{\trajH}^{(i)}, \bm{x}_{\trajH}^{(i)})}\leq\frac{B}{N}\sqrt{\operatorname{trace}\left(T_{\bm{K}}^\scale{N}\right)}
    .
    \end{align*}
    Note that the different exponent in $n$ stems from the different definitions of the operator and matrix $\operatorname{trace}$.
\end{Proof}
Apart from the data density dependencies, the complexity of the hypothesis space is captured by the trace of the integral operator, the Grammian, iterating on a well-known property of RKHS methods. Naturally, this provides little insight asymptotically as the trace of an operator is not immediately assessable. Treatment of the trace in the asymptotic case is provided in the following result on the excess risk of KKR, which we are now ready to state.
\GeneralizationGapOfKKR*
\begin{Proof}[Theorem~\ref{theorem:generalizationRiskKKR}]\label{proof:theorem:generalizationRiskKKR}
    The statements follow by combining Theorem~\ref{theorem:RademacherGeneralizationRiskBound} with approximations of the Rademacher complexities of the Koopman kernel RKHS provided in Lemma~\ref{lemma:RademacherComplexitiesKK}. In the asymptotic case, the behaviour of  $\operatorname{trace}\left(T_{\bm{K}}\right)$ is of interest. We employ the following upper bound.
    \begin{align*}
    \operatorname{trace}\left(T_{\bm{K}}\right) 
    &=
    \sum_{i}^{} \left\langle T_{\bm{K}} e_i,  e_i \right\rangle
    & \textit{by definition}
    \\
    &=
    \sum_{i}^{} \left\langle T_{\bm{K}}^{\frac{1}{2}} e_i,  \adjoint{T_{\bm{K}}^{\frac{1}{2}}} e_i \right\rangle
    & \textit{trace-class property}
    \\
    &=
    \normalint_{\mathbb{X}} \left\langle \bm{K}(\cdot, \bm{x}_{\trajH}),  \bm{K}(\cdot , \bm{x}_{\trajH}) \right\rangle\d{\bm{x}_{\trajH}}
    & \textit{kernel trick}
    \\
    &=
    \normalint_{\Set{X}}^{} \bm{K}(\bm{x}_{\trajH}, \bm{x}_{\trajH}) \d{\bm{x}_{\trajH}}
    & \textit{reproducing property}
    \\
    &= 
    \normalint_{\Set{X}}^{} \normalint_{\rho_{\deig}}\bm{K}^{\deig}(\bm{x}_{\trajH}, \bm{x}_{\trajH}) \d{\deig}\d{\bm{x}_{\trajH}}
    & \textit{Koopman kernel}
    \\
    &= 
    \normalint_{\Set{X}}^{} \normalint_{\rho_{\deig}}\bm{C}(\deig, H) \bm{K}\naught^{\deig}(\bm{x}_{\trajH}, \bm{x}_{\trajH}) \d{\deig}\d{\bm{x}_{\trajH}}
    & \textit{Koopman kernel flow}
    \\
    &\leq
    \|\bm{C}(\deig, H)\|\normalint_{\Set{X}}^{} \normalint_{\rho_{\deig}}\bm{K}^{\deig}(\bm{x}_{\trajH}, \bm{x}_{\trajH}) \d{\deig}\d{\bm{x}_{\trajH}}
    & \textit{Fubini's Theorem}
    \\
    &\leq
    \|\bm{C}(\deig, H)\|\sup\limits_{\bm{x}_{\trajH}} [\bm{K}_0^{\mu}] H\normalint_{\Set{X}}^{}\normalint_{\rho_{\deig}} \d{\bm{x}_{\trajH}} \d{\deig }
    & \textit{Gershgorin Circle Theorem}   
    \\
    &=
    \|\bm{C}(\deig, H)\|\kappa H\normalint_{\Set{X}}^{}\normalint_{\rho_{\deig}} \d{\bm{x}} \d{\deig }
    & \textit{bounded kernel}    
    \\
    &=
    \|\bm{C}(\deig, H)\| \kappa H {\color{white}\normalint_{\Set{X}}^{} \d{\bm{x}}}
    & \textit{appropriate normalization}
    \\
    &\leq
    1H \kappa H=\kappa H^2 {\color{white}\normalint_{\Set{X}}^{} \d{\bm{x}}}
    & \textit{Gershgorin Circle Theorem (again)}    
    \end{align*}
    Where $\bm{K}^{\deig}=\bm{C}(\deig, H) \bm{K}\naught^{\deig}$ is the decomposition of the eigenfunction kernel into an evaluation at a point \emph{in space} $\bm{K}\naught^{\deig}=\evalat{\bm{K}^{\deig}}{t=0}$ and its flow \emph {in time} $\bm{C}(\deig, H)=\deig^{ k}\otimes \adjoint{\deig^{k}}\in\Complex^{H\times H}$ defined by the outer product of the eigenfunction flow. Consequently, the last inequality follows from the fact that exponential frequencies, especially when sampled from the unit disk, do not explode within a finite number of steps $H$. \par%
    The last ingredient we need is an approximation of the Lipschitz constant $L(\loss\naught)$. Consider the Representation-Error $\|\bm{y}_{\trajT} - \hat{M}(\bm{x}_{\trajT})\|\leq R$. On our non-recurrent domain of \textit{finite} time $\bm{y}_{\trajT}$ does not diverge, neither does $\hat{M}(\bm{x}_{\trajT})$, since we solve a regularized problem. This entails the boundedness of $\loss$ by $R$. Thus, the squared error loss is Lipschitz with constant $L=\sup_{\bm{x}}\frac{\partial}{\partial \bm{x}}\loss(\bm{x}) =2R$.
    \par%
    We can now combine the preceding investigations with
    Theorem~\ref{theorem:RademacherGeneralizationRiskBound}
    and obtain our claim immediately.
\end{Proof}

\section{Numerical Evaluation Details and Additional Experiments}\label{supl:exp}
All of the experiments were performed on a machine with {2TB} of RAM, 8 NVIDIA Tesla P100 16GB GPUs and 4 AMD EPYC 7542 CPUs.

The comparisons to PCR (EDMD) and RRR are done utilizing MIT-licensed code accompanying \cite{Kostic2022LearningSpaces} available at {\urlstyle{rm}\url{ https://github.com/csml-iit-ucl/kooplearn}\footnote{last accessed version {\ttfamily "0.1.24"} at \url{https://github.com/csml-iit-ucl/kooplearn/tree/legacy_kooplearn} from  April 25, 2023}}. Signature kernels implementation is that of Sig-PDEs accompanying \cite{Salvi2021ThePDE}, available at {\urlstyle{rm}\url{ https://github.com/crispitagorico/sigkernel}}\footnote{last accessed version from  July 25, 2023}. For forecasting with Sig-PDE we fit a ridge regressor from observation time-delays and times to their successor. The prediction is then concatenated to the history and used to forecast subsequent steps. To ensure that Sig-PDE forecasts the same times in $\{0, \dots, H\Delta t \}$ we simulate the systems backwards in time and train Sig-PDE with observations from the interval $\{-l\Delta t , \dots, H\Delta t\}$. 

\subsection{Numerical Evaluation Details}
{\bf Normalizing the invariance transform~}
We normalize the invariance transformation of each eigenvalue by the norm of its pullback $\|\exp{-\eig t}\|_{\Set{T}}$/ $\|\deig^{h}\|_{\Set{H}}$. Normalizing increases numerical stability significantly as for discrete-time eigenvalues close to the origin the pullback $\deig^{-k }$ go to infinity. Beyond mere numerical convenience, this also provides intuition on what the invariance transformation does. Consider the aforementioned case $\mu\rightarrow 0$, then the eigenfunction decays infinitesimally fast: the invariance transformation becomes an indicator at the final time $\delta_T(t)$.

{\bf Details on the bi-stable system experiment}
We chose $N=50$ datapoints. For the base kernel we utilize the radial basis function (RBF) kernel $k(\bm{x},\bm{x}^\prime)=\operatorname{e}^{\frac{1}{2 \ell^2}\|\bm{x}-\bm{x}^\prime\|^2}$ with a length scale of $\ell=0.05$, covering the whole state space, while allowing for sufficient distinction of trajectories due the time-horizon $H=14$ fulfilling our non-recurrence assumption. We trained models for EDMD and KKR with predictor rank $D$ in a range from 1 to 100 and chose the best performing for each method. Unsurprisingly, KKR performs best with 100 eigenfunctions while EDMD attains its minimizer at 10. 

{\bf Van der Pol oscillator experiment detail}
We utilize RBF kernels with a length scale of $\ell=0.1$.

\subsection{Additional Experiments}
 {\bf Eigenspace and sample cardinality dependence}%
\input{plots/appendix_bistable_risk}%
\input{plots/appendix_vanderpol_risk}%
To provide more intuition on how our method, and as a baseline EDMD, performs dependent on the number of samples and eigenfunctions used, we provide parameterized versions of the experiments from the main text. Bite that the bi-stable system experiment is here run with parameters $a=4$, $b=-16$. Figure~\ref{fig:appendixBistableNandDo} depicts these dependencies for the bi-stable system, while Figure~\ref{fig:appendixVdPNandDo} displays the same experiments for the Van der Pol oscillator. We observe that KKR admits the sane property of increased excess and test performance with increasing cardinality of eigenspaces $D$. It also becomes clear that, due to limited data, increase in the number of eigenfunctions has, at some point, diminished returns for the test risk of KKR. Nevertheless, additional eigenfunctions do not deteriorate the test risk, a salient feature or our approach compared to EDMD that might yield worse performance on test data -- as predicted by~\cite{Kostic2022LearningSpaces}.
 
{\bf Validation of other theoretical results~}
Using Monte-Carlo-Integration, we verify the convergence of the kernel \eqref{eq:KEIGKernTrajDT2} in the misspecified case by Figure~\ref{figure:ConvergenceToDistrinutionKernel}. We sample eigenvalues from the uniform distribution on the complex unit disk. We use the kernel with $D=2\times 10^5$ as a baseline and average the difference of the operator-valued kernel to the baseline with the Frobenius norm. Results are averaged over $N=5$ different points over 20 (i.i.d.) runs each with time-horizon $H=14$.
\input{plots/appendix_kernel_convergence}

{\bf Kármán vortex street}%
\input{plots/appendix_karman_signal}%
\input{plots/appendix_karman_experiment}%
In fluid dynamics, a Kármán vortex street is a phenomenon that is observed when a laminar flow is disturbed by a solid object. We consider a cylinder. After a settling phase, the transient, periodically oscillating vortices behind the cylinder eventuate. This phenomenon occurs, for example, in the airflow behind a car or a wind turbine. Therefore, predicting the effect of vortex streets on velocity fields is highly relevant for engineers in the aero- and hydro-dynamic design of systems since the frequency of oscillation might cause undesirable resonance. Fluid dynamics simulations solving some variation of the Navier-Stokes equations, usually by discretizing space into a grid, are employed to predict the aforementioned effects. However, integrating these simulations in complex multi-physics simulations is challenging due to their relatively high computational complexity -- making fluid simulation a bottleneck. Thus, surrogate modelling of the effect of interest through a faster-to-evaluate model is of great interest. Nevertheless, as the states of a fluid simulation are usually velocities or other quantities at each grid point, the data available to train surrogate models is high-dimensional and, thus, often challenging to handle. 

To demonstrate that our method is capable of performing well with high dimensional data in the context described above, we employ it to obtain a simplified representation -- an \textit{LTI predictor} -- of the measurements of a sensor in a Kármán vortex street under variation of the initial condition. The variation is a deviation in the cylinder placement. The setup is depicted in Figure~\ref{figure:cylinderFlow}. To obtain the ground truth, we employ a solver based on the Lattice-Boltzmann 
Method \cite{LBM} from an MIT-licensed implementation available at {\urlstyle{rm}\url{  https://github.com/Ceyron/machine-learning-and-simulation/tree/main/english/simulation_scripts}}. We specify a Reynolds number of $40$, a $100 \times50 $ grid and an inlet velocity at $(0, y)$ of $0.05\nicefrac{m}{s}$ in $x$-direction. The cylinder position is varied by up to three grid points in each direction around $(20, 25)$, amounting to 49 different initial conditions, for which sample trajectories are computed. We randomly split those into 44 training and five testing samples. 
Simulation yields our state -- the velocity magnitudes at each grid point $d=100\times 50=5000$ -- over horizon length $H=99$. Therefore, a  trajectory can be interpreted as a sequence of images. A sample trajectory can be found next to this document in the supplemental. We place a virtual sensor at (80, 25), such that the corresponding velocity magnitude is our observable. Using the knowledge that the Kármán vortex street admits stable periodic behaviour, we select Koopman operator eigenvalues $\eig$ that are purely imaginary, for the stable periodic manifold, or purely decaying, for the transient regime \cite{KoopBook,Mezic2020SpectrumGeometry}: $\deig = \operatorname{e}^{\eig \Delta t}$, where $\eig \sim \rho_{\lambda}=\operatorname{uniform}\left(\{\pm aj, -a|0\leq a\leq1\}\right)$. We fit a KKR model with $D=500$ and an RBF base kernel with length scale $\ell=30$. The model enables us to forecast the observable using an image of the velocity magnitudes -- a 5000 dimensional vector -- as input. In Figure~\ref{figure:KarmanTrajectories}, our model's prediction is compared to ground truth. We observe that training trajectories are accurately reconstructed, with good performance on test data, despite the low number of training samples $N=45$. Notably, reproducing the dataset using KKR takes $\approx0.05$ seconds (average over 1000 calls), while simulating the ground tooth takes $\approx1$ second per run (average over 49 runs), both using one GPU unit -- demonstrating suitability for surrogate models.

\fi

\end{document}

%% file: plots/risk_van_der_pol.tex
%
{
\DeclareRobustCommand\sampleline[1]{%
  \tikz\draw[#1] (0,0) (0,\the\dimexpr\fontdimen22\textfont2\relax)
  -- (10pt,\the\dimexpr\fontdimen22\textfont2\relax);%
}
\pgfplotsset{ticklabel style = {font=\scriptsize\sffamily},
	every axis label/.append style={font=\small\sffamily, yshift=6pt},
	legend style = {font=\scriptsize\sffamily},title style={yshift=-7pt, font = \small\sffamily} }
\begin{figure}[!tpb]
\centering
        \begin{tikzpicture}
            \renewcommand{\pathtoresults}[0]{plots/data_van_der_pol/test_risk_over_N_by_D}
                \begin{loglogaxis}[
                        name=plot1,
                        height = 3.5cm,
                        width=0.515\textwidth,
                        xmin=8, xmax=200,
                        xlabel={\scriptsize $N$},
                        xmajorgrids,
                        ymin=5e-3, ymax=1e0,
                        ylabel={\scriptsize Generalization Gap},
                        yticklabel style={xshift=0pt},
                        ylabel style={yshift=-100pt},
                        ylabel near ticks, yticklabel pos=left,
                        ymajorgrids,
                        legend columns=2,
                        legend style={
                        nodes={scale=0.75, transform shape},
                        anchor=north, xshift=-0.1cm},
                        legend pos=south west,
                        title={\scriptsize Training Sample Cardinality Dependence},
                        title style = {xshift=-1ex,yshift=2.5ex},
                         ylabel style={xshift=-10pt},
                    ]
                    \addlegendimage{blue, thick}
                    \addlegendentry{$\!$Ours};
                    \addlegendimage{red, thick}
                    \addlegendentry{$\!$RRR~
                    };
                    \addlegendimage{green, thick}
                    \addlegendentry{$\!$Sig-PDE~
                    };
                    \addlegendimage{orange, thick}
                    \addlegendentry{$\!$PCR~
                    };
                    \addlegendimage{black, thick, dashed}
                    \addlegendentry{$\!$$\mathcal{O}(\nicefrac{1}{\sqrt{N}})$};
                    \addriskplot{500}{KKR}{\pathtoresults}{blue}{blue}{solid, thick}{}
                    \addriskplot{100}{RRR}{\pathtoresults}{red}{red}{solid, thick}{}
                    \addriskplot{62}{PCR}{\pathtoresults}{orange}{orange}{solid, thick}{}
                    \addriskplot{10}{SK}{\pathtoresults}{green}{green}{solid, thick}{}
                    \draw[black, thick, dashed,scale=0.5,domain=2:200,smooth,variable=\x]
        plot ({\x},{-0.5*\x+1.0});
                \end{loglogaxis}
                \renewcommand{\pathtoresults}[0]{plots/data_van_der_pol/test_risk_over_D_by_N}
                \begin{loglogaxis}[
                        name=plot2,
                        at=(plot1.east),
                        xshift=0.1cm,
                        anchor=west,
                        height = 3.5cm,
                        width=0.515\textwidth,
                        xmin=8, xmax=200,
                        xlabel={},
                        xmajorgrids,
                        axis x line*=bottom,
                        xmax=1e2,
                        ymin=5e-3, ymax=1e0,
                        ylabel={\scriptsize Test Risk},
                        ymajorgrids,yticklabel pos=right,ylabel near ticks, 
                        legend columns=3,
                        legend style={column sep=1pt, text opacity=1,draw opacity=1,fill opacity=0.2, at={(0.555,1.0)},anchor=north},
                        title={\scriptsize Eigenspace Cardinality $D$},
                         title style={at={(0.5,0.1)},anchor=north,yshift=-0ex},
                          ylabel style={xshift=-10pt},
                    ]
                    \addriskplot{200}{KKR}{\pathtoresults}{blue}{blue}{solid, thick}{}
                    \addriskplot{200}{PCR}{\pathtoresults}{orange}{orange}{solid, thick}{}
                    \addriskplot{200}{RRR}{\pathtoresults}{red}{red}{solid, thick}{}
                \end{loglogaxis}
                \begin{semilogyaxis}[
                        name=plot3,
                        at=(plot1.east),
                        xshift=0.1cm,
                        anchor=west,
                        height = 3.5cm,
                        width=0.515\textwidth,
                        xmin=2, xmax=10,
                        axis x line*=top,
                        xtick={2, 4, 6, 8, 10},
                        xlabel={},
                        xlabel near ticks,
                        xmajorgrids,
                        ymin=5e-3, ymax=1e0,
                        ylabel={\scriptsize },
                        ymajorgrids,yticklabel pos=right,ylabel near ticks, 
                        legend columns=3,
                        legend style={column sep=1pt, text opacity=1,draw opacity=1,fill opacity=0.2, at={(0.665,1.0)},anchor=north},
                        title={\scriptsize Delay Length $l$},
                        title style={yshift=2.5ex,},
                    ]
                    \addriskplot{200}{SK}{\pathtoresults}{green}{green}{solid, thick}{}
                \end{semilogyaxis}
        \end{tikzpicture}

\vspace{-0.33cm}
    \caption{
    Forecasting risks (20  i.i.d. runs) for the Van der Pol system over a time-horizon $H=14$ ($T=1s$). \textbf{Left:} Generalization gap for the best $D$ / $l$ ({ours} 500, {PCR} 62, {RRR} 100, {RR-Sig-PDE} 10) is depicted with a growing number of data points. \textbf{Right:} Test risk behavior with an increasing amount of eigenspaces is shown for $N=200$. Shaded areas depict min-max risk intervals.}
\label{fig:NandDo}
\vspace{-0.75\intextsep}
\end{figure}
}

%% file: plots/risk_karman_hp.tex
%
{
\DeclareRobustCommand\sampleline[1]{%
  \tikz\draw[#1] (0,0) (0,\the\dimexpr\fontdimen22\textfont2\relax)
  -- (10pt,\the\dimexpr\fontdimen22\textfont2\relax);%
}
\pgfplotsset{ticklabel style = {font=\scriptsize\sffamily},
	every axis label/.append style={font=\small\sffamily, yshift=6pt},
	legend style = {font=\scriptsize\sffamily},title style={yshift=-7pt, font = \small\sffamily} }
\begin{figure}[!tpb]
\centering
        \begin{tikzpicture}
            \renewcommand{\pathtoresults}[0]{plots/data/karman/hyperparameter_sensitivities/}
                \begin{axis}[
                        name=plot1,
                        height = 3.5cm,
                        width=0.515\textwidth,
                        xmin=-1e0, xmax=98,
                        xmajorgrids,
                        ymin=-3e0, ymax=6e1,
                        ylabel={\scriptsize $\sum$ Absolute Error},
                        yticklabel style={xshift=0pt},
                        ylabel near ticks, yticklabel pos=left,
                        ymajorgrids,
                        legend columns=2,
                        legend style={
                        nodes={scale=0.75, transform shape},
                        anchor=north, xshift=0.5cm},
                        legend pos=north west,
                        title={\scriptsize  timesteps},
                        title style={at={(0.5,0)},anchor=north,yshift=0ex},
                         ylabel style={xshift=-10pt},
                    ]
                    \addlegendimage{blue!66!yellow, thick}
                    \addlegendentry{$\!$Ours train};
                    \addlegendimage{blue, thick}
                    \addlegendentry{$\!$Ours test};
                    \addlegendimage{yellow!20!red, thick}
                    \addlegendentry{$\!$PCR train~
                    };
                    \addlegendimage{orange, thick}
                    \addlegendentry{$\!$PCR test~
                    };
                    \addriskplot{200}{KKR}{\pathtoresults train_cum_risk_karman_D}{blue!66!yellow}{blue!66!yellow}{solid, thick}{}
                    \addriskplot{200}{KKR}{\pathtoresults 
                    test_cum_risk_karman_D}{blue}{blue}{solid, thick}{}
                    \addriskplot{200}{PCR}{\pathtoresults train_cum_risk_karman_D}{orange}{orange}{solid, thick}{}
                    \addriskplot{200}{PCR}{\pathtoresults test_cum_risk_karman_D}{yellow!20!red}{yellow!20!red}{solid, thick}{}
                \end{axis}
            \renewcommand{\pathtoresults}[0]{plots/data/karman/hyperparameter_sensitivities/}
                \begin{loglogaxis}[
                        name=plot2,
                        at=(plot1.east),
                        xshift=0.1cm,
                        anchor=west,
                        height = 3.5cm,
                        width=0.515\textwidth,
                        xmin=8, xmax=200,
                        xlabel={},
                        xmajorgrids,
                        xmin=5e-0, xmax=1e4,
                        ymin=9e-3, ymax=30e0,
                        ylabel={\scriptsize Risk},
                        ymajorgrids,yticklabel pos=right,ylabel near ticks, 
                        legend columns=3,
                        legend style={column sep=1pt, text opacity=1,draw opacity=1,fill opacity=0.2, at={(0.555,1.0)},anchor=north},
                        title={\scriptsize RBF kernel lengthscale $\ell$},
                         title style={at={(0.5,0)},anchor=north,yshift=-0ex},
                         ylabel style={xshift=-10pt},
                    ]
                    \addriskplot{200}{KKR}{\pathtoresults train_risk_karman_D}{blue!66!yellow}{blue!66!yellow}{solid, thick}{}
                    \addriskplot{200}{KKR}{\pathtoresults 
                    test_risk_karman_D}{blue}{blue}{solid, thick}{}
                    \addriskplot{200}{PCR}{\pathtoresults train_risk_karman_D}{orange}{orange}{solid, thick}{}
                    \addriskplot{200}{PCR}{\pathtoresults test_risk_karman_D}{yellow!20!red}{yellow!20!red}{solid, thick}{}
                \end{loglogaxis}
        \end{tikzpicture}
\vspace{-0.33cm}
\caption{Cumulative error and forecast risks (5 train-test~splits) for flow past cylinder data and $H=99$. Our KKR with orders-of-magnitude greater usable $\ell$-range and accuracy. \textbf{Left:} Cumulative absolute error for the best $D/ \ell$ ({ours} 200/70, {PCR} 200/35) is depicted over timesteps. \textbf{Right:} Forecast risk for 99 steps within a range of RBF lengthscales. Shaded areas depict min-max intervals.}
\label{fig:accAndLen}
\vspace{-0.5\intextsep}
\end{figure}
}

%% file: plots/appendix_bistable_risk.tex
%
{
\DeclareRobustCommand\sampleline[1]{%
  \tikz\draw[#1] (0,0) (0,\the\dimexpr\fontdimen22\textfont2\relax)
  -- (10pt,\the\dimexpr\fontdimen22\textfont2\relax);%
}
\pgfplotsset{ticklabel style = {font=\scriptsize\sffamily},
	every axis label/.append style={font=\small\sffamily, yshift=6pt},
	legend style = {font=\scriptsize\sffamily},title style={yshift=-0pt, font = \small\sffamily} }
\begin{figure}
\centering
\begin{tikzpicture}
    \renewcommand{\pathtoresults}[0]{plots/data/bistable/overN/excess_risk_over_N_by_D}
        \begin{loglogaxis}[
                name=plot1,
                height = 6cm,
                width=0.515\textwidth,
                xmin=8, xmax=200,
                xlabel={$N$},
                xmajorgrids,
                ymin=1e-5, ymax=9e0,
                ylabel={Generalization Gap},
                yticklabel style={xshift=0pt},
                ylabel style={yshift=-10pt},
                ylabel near ticks, yticklabel pos=left,
                ymajorgrids,
                legend columns=3,
                legend style={
                anchor=north},
                legend pos=south west,
                title={Training Sample Cardinality Dependence},
            ]
            \addlegendimage{blue, thick}
            \addlegendentry{Ours};
            \addlegendimage{red, thick}
            \addlegendentry{EDMD~\cite{Kostic2022LearningSpaces}};
            \addlegendimage{black, thick, dashed}
            \addlegendentry{$\mathcal{O}(\nicefrac{1}{\sqrt{N}})$};
            \addriskplot{10}{ours_isometric_}{\pathtoresults}{blue}{blue}{dotted, thick}{}
            \addriskplot{10}{theirs}{\pathtoresults}{red}{red}{dotted, thick}{}
            \addriskplot{41}{ours_isometric_}{\pathtoresults}{blue}{blue}{dashed, thick}{}
            \addriskplot{41}{theirs}{\pathtoresults}{red}{red}{dashed, thick}{}
            \addriskplot{400}{ours_isometric_}{\pathtoresults}{blue}{blue}{}{}
            \addriskplot{400}{theirs}{\pathtoresults}{red}{red}{}{}
            \draw[black, thick, dashed,scale=0.5,domain=2:200,smooth,variable=\x]
plot ({\x},{-0.5*\x+1.0});
        \end{loglogaxis}
        \renewcommand{\pathtoresults}[0]{plots/data/bistable/overD//test_risk_over_D_by_N}
        \begin{loglogaxis}[
                name=plot2,
                at=(plot1.east),
                xshift=0.1cm,
                anchor=west,
                height = 6cm,
                width=0.515\textwidth,
                xmin=8, xmax=200,
                xlabel={$D$},
                xmajorgrids,
                ymin=1e-5, ymax=9e0,
                ylabel={Test Risk},
                ymajorgrids,yticklabel pos=right,ylabel near ticks, 
                legend columns=3,
                legend style={column sep=1pt, text opacity=1,draw opacity=1,fill opacity=0.2, at={(0.665,1.0)},anchor=north},
                title={Eigenspace Cardinality Dependence}
            ]
            \addriskplot{19}{ours_isometric_}{\pathtoresults}{blue}{blue}{dotted, thick}{}
            \addriskplot{19}{theirs}{\pathtoresults}{red}{red}{dotted, thick}{}
            \addriskplot{62}{ours_isometric_}{\pathtoresults}{blue}{blue}{dashed, thick}{}
            \addriskplot{62}{theirs}{\pathtoresults}{red}{red}{dashed, thick}{}
            \addriskplot{200}{ours_isometric_}{\pathtoresults}{blue}{blue}{}{}
            \addriskplot{200}{theirs}{\pathtoresults}{red}{red}{}{}
            
        \end{loglogaxis}
\end{tikzpicture}
\vspace{-0.2cm}
    \caption{
    Forecasting risks for the bi-stable system over a time-horizon $H=14$. \textbf{Left:} Forecast generalization gap for $D\in\{10:\text{\sampleline{dotted}}, 41:\text{\sampleline{dashed}}, 400:\text{\sampleline{}}\}$  is depicted with a growing number of data points. \textbf{Right:} Test risk behavior with an increasing amount of eigenspaces is shown for $N\in\{19:\text{\sampleline{dotted}}, 62:\text{\sampleline{dashed}},200:\text{\sampleline{}}\}$, demonstrating the benefits of KKR.}
\label{fig:appendixBistableNandDo}
\end{figure}
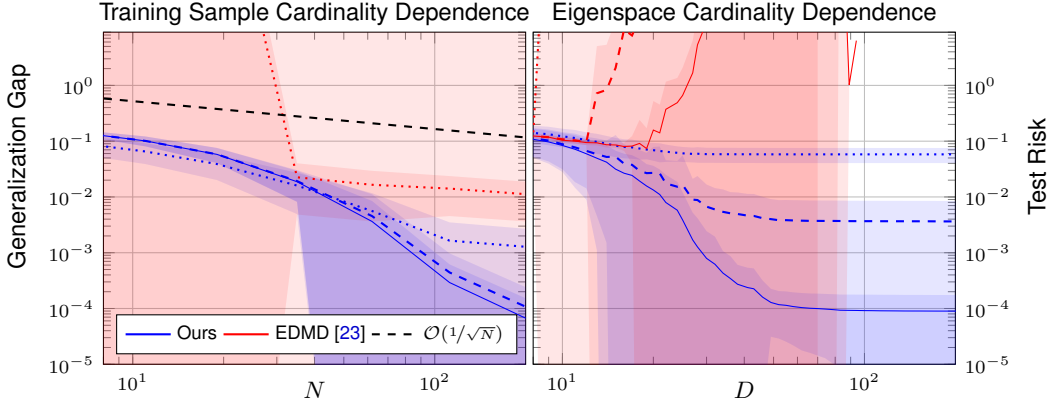}

%% file: plots/appendix_vanderpol_risk.tex
%
{
\DeclareRobustCommand\sampleline[1]{%
  \tikz\draw[#1] (0,0) (0,\the\dimexpr\fontdimen22\textfont2\relax)
  -- (10pt,\the\dimexpr\fontdimen22\textfont2\relax);%
}
\pgfplotsset{ticklabel style = {font=\scriptsize\sffamily},
	every axis label/.append style={font=\small\sffamily, yshift=6pt},
	legend style = {font=\scriptsize\sffamily},title style={yshift=0pt, font = \small\sffamily} }
\begin{figure}
\centering
\begin{tikzpicture}
    \renewcommand{\pathtoresults}[0]{plots/data/vanderpol/overN/excess_risk_over_N_by_D}
        \begin{loglogaxis}[
                name=plot1,
                height = 6cm,
                width=0.515\textwidth,
                xmin=8, xmax=200,
                xlabel={$N$},
                xmajorgrids,
                ymin=1e-3, ymax=9e0,
                ylabel={Generalization Gap},
                yticklabel style={xshift=0pt},
                ylabel style={yshift=-10pt},
                ylabel near ticks, yticklabel pos=left,
                ymajorgrids,
                legend columns=3,
                legend style={
                anchor=north},
                legend pos=south west,
                title={Training Sample Cardinality Dependence},
            ]
            \addlegendimage{blue, thick}
            \addlegendentry{Ours};
            \addlegendimage{red, thick}
            \addlegendentry{EDMD~\cite{Kostic2022LearningSpaces}};
            \addlegendimage{black, thick, dashed}
            \addlegendentry{$\mathcal{O}(\nicefrac{1}{\sqrt{N}})$};
            \addriskplot{10}{ours_isometric_}{\pathtoresults}{blue}{blue}{dotted, thick}{}
            \addriskplot{10}{theirs}{\pathtoresults}{red}{red}{dotted, thick}{}
            \addriskplot{50}{ours_isometric_}{\pathtoresults}{blue}{blue}{dashed, thick}{}
            \addriskplot{50}{theirs}{\pathtoresults}{red}{red}{dashed, thick}{}
            \addriskplot{200}{ours_isometric_}{\pathtoresults}{blue}{blue}{}{}
            \addriskplot{200}{theirs}{\pathtoresults}{red}{red}{}{}
            \draw[black, thick, dashed,scale=0.5,domain=2:200,smooth,variable=\x]
plot ({\x},{-0.5*\x+1.0});
        \end{loglogaxis}
        \renewcommand{\pathtoresults}[0]{plots/data/vanderpol/overD//test_risk_over_D_by_N}
        \begin{loglogaxis}[
                name=plot2,
                at=(plot1.east),
                xshift=0.1cm,
                anchor=west,
                height = 6cm,
                width=0.515\textwidth,
                xmin=8, xmax=200,
                xlabel={$D$},
                xmajorgrids,
                ymin=1e-3, ymax=9e0,
                ylabel={Test Risk},
                ymajorgrids,yticklabel pos=right,ylabel near ticks, 
                legend columns=3,
                legend style={column sep=1pt, text opacity=1,draw opacity=1,fill opacity=0.2, at={(0.665,1.0)},anchor=north},
                title={Eigenspace Cardinality Dependence}
            ]
            \addriskplot{19}{ours_isometric_}{\pathtoresults}{blue}{blue}{dotted, thick}{}
            \addriskplot{19}{theirs}{\pathtoresults}{red}{red}{dotted, thick}{}
            \addriskplot{62}{ours_isometric_}{\pathtoresults}{blue}{blue}{dashed, thick}{}
            \addriskplot{62}{theirs}{\pathtoresults}{red}{red}{dashed, thick}{}
            \addriskplot{200}{ours_isometric_}{\pathtoresults}{blue}{blue}{}{}
            \addriskplot{200}{theirs}{\pathtoresults}{red}{red}{}{}
            
        \end{loglogaxis}
\end{tikzpicture}
\vspace{-0.2cm}
    \caption{
    Forecasting risks for the Van der Pol oscillator over a time-horizon $H=14$. \textbf{Left:} Forecast generalization gap for $D\in\{10:\text{\sampleline{dotted}}, 50:\text{\sampleline{dashed}}, 200:\text{\sampleline{}}\}$ is depicted with a growing number of data points. \textbf{Right:} Test risk behavior with an increasing amount of eigenspaces is shown for $N\in\{19:\text{\sampleline{dotted}}, 62:\text{\sampleline{dashed}},200:\text{\sampleline{}}\}$, demonstrating the benefits of KKR.}
\label{fig:appendixVdPNandDo}
\end{figure}}

%% file: plots/appendix_kernel_convergence.tex
%
\DeclareRobustCommand\sampleline[1]{%
  \tikz\draw[#1] (0,0) (0,\the\dimexpr\fontdimen22\textfont2\relax)
  -- (10pt,\the\dimexpr\fontdimen22\textfont2\relax);%
}
\pgfplotsset{ticklabel style = {font=\scriptsize\sffamily},
	every axis label/.append style={font=\small\sffamily, yshift=6pt},
	legend style = {font=\scriptsize\sffamily},title style={yshift=-0pt, font = \small\sffamily} }
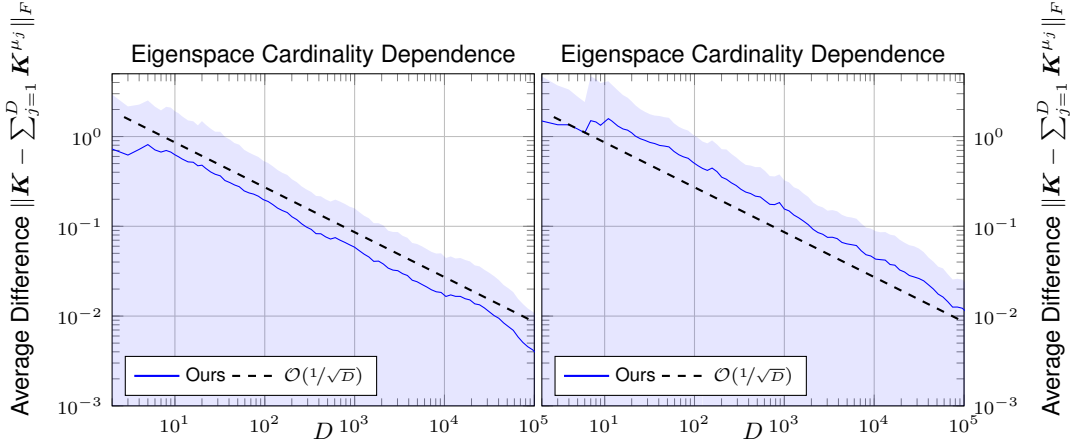
\begin{figure}
\centering
\begin{tikzpicture}
    \renewcommand{\pathtoresults}[0]{plots/data/vanderpol/}
        \begin{loglogaxis}[
                name=plot1,
                height = 6cm,
                width=0.515\textwidth,
                xmin=2e0, xmax=1e5,
                xlabel={$D$},
                xmajorgrids,
                ymin=1e-3, ymax=5e0,
                ylabel={Average Difference $\|\bm{K}-\textstyle\sum^D_{j=1}{\bm{K}^{\deig_j}}\|_{F}$},
                yticklabel style={xshift=0pt},
                ylabel style={yshift=-10pt},
                ylabel near ticks, yticklabel pos=left,
                ymajorgrids,
                legend columns=3,
                legend style={
                anchor=North},
                legend pos=south west,
                title={Eigenspace Cardinality Dependence},
            ]
            \addlegendimage{blue, thick}
            \addlegendentry{Ours};
            \addlegendimage{black, thick, dashed}
            \addlegendentry{$\mathcal{O}(\nicefrac{1}{\sqrt{D}})$};
            \addplot[color=green, name path=path1, color=blue] table[x=D,y=M]{\pathtoresults kernel_error_frob_norm_R20_D200000_N5_vanderpol.csv};
            \addplot[draw=none, name path=D]
            table[x=D,y=L]{\pathtoresults kernel_error_frob_norm_R20_D200000_N5_vanderpol.csv};
            \addplot[draw=none, name path=U]
            table[x=D,y=U]{\pathtoresults kernel_error_frob_norm_R20_D200000_N5_vanderpol.csv};
            \addplot[fill=blue, fill opacity=0.1] fill between [of=D and U];
            \draw[black, thick, dashed,scale=0.5,domain=2:200,smooth,variable=\x]
plot ({\x},{-0.5*\x+2.0});
        \end{loglogaxis}
    \renewcommand{\pathtoresults}[0]{plots/data/bernoulli3/}
        \begin{loglogaxis}[
                name=plot2,
                at=(plot1.east),
                xshift=0.1cm,
                anchor=west,
                height = 6cm,
                width=0.515\textwidth,
                xmin=2e0, xmax=1e5,
                xlabel={$D$},
                xmajorgrids,
                ymin=1e-3, ymax=5e0,
                ylabel={Average Difference $\|\bm{K}-\textstyle\sum^D_{j=1}{\bm{K}^{\deig_j}}\|_{F}$},
                ymajorgrids,yticklabel pos=right,ylabel near ticks, 
                legend columns=3,
                legend pos=south west,
                title={Eigenspace Cardinality Dependence}
            ]
            \addlegendimage{blue, thick}
            \addlegendentry{Ours};
            \addlegendimage{black, thick, dashed}
            \addlegendentry{$\mathcal{O}(\nicefrac{1}{\sqrt{D}})$};
            \addplot[color=green, name path=path1, color=blue] table[x=D,y=M]{\pathtoresults kernel_error_frob_norm_R20_D200000_N5.csv};
            \addplot[draw=none, name path=D]
            table[x=D,y=L]{\pathtoresults kernel_error_frob_norm_R20_D200000_N5.csv};
            \addplot[draw=none, name path=U]
            table[x=D,y=U]{\pathtoresults kernel_error_frob_norm_R20_D200000_N5.csv};
            \addplot[fill=blue, fill opacity=0.1] fill between [of=D and U];
            \draw[black, thick, dashed,scale=0.5,domain=2:200,smooth,variable=\x]
plot ({\x},{-0.5*\x+2.0});
        \end{loglogaxis}
\end{tikzpicture}
\vspace{-0.2cm}
    \caption{
    Norm difference of the sampled kernel to the specified kernel. \textbf{Left:} Norm difference of the kernel for the Van der Pol oscillator is depicted with a growing number of eigenvalues. \textbf{Right:} Norm difference of the kernel for the bi-stable system is depicted with a growing number of eigenvalues. 
    }
\label{figure:ConvergenceToDistrinutionKernel}
\end{figure}


%% file: plots/appendix_karman_signal.tex
{
\pgfplotsset{ticklabel style = {font=\scriptsize\sffamily},
	every axis label/.append style={font=\small\sffamily, yshift=6pt},
	legend style = {font=\scriptsize\sffamily},title style={yshift=-0pt, font = \small\sffamily} }
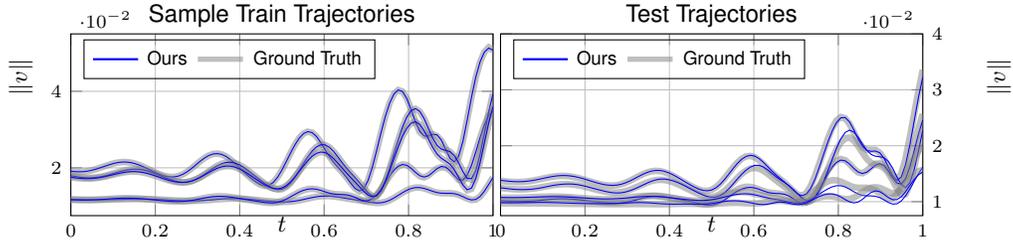
\begin{figure}
\centering
\begin{tikzpicture}
\renewcommand{\pathtoresults}[0]{plots/data/karman/trajectories/}
        \begin{axis}[
                name=plot1,
                height = 4cm,
                width=0.515\textwidth,
                xmin=0, xmax=1,
                xlabel={$t$},
                xmajorgrids,
                ymin=0.0075, ymax=.055,
                ylabel={$\|v\|$},
                yticklabel style={xshift=0pt},
                ylabel style={yshift=-10pt},
                ylabel near ticks, yticklabel pos=left,
                ymajorgrids,
                legend columns=3,
                legend style={column sep=1pt, text opacity=1,draw opacity=1,fill opacity=0.2,anchor=north west},
                legend pos=north west,
                title={Sample Train Trajectories},
            ]
            \addlegendimage{blue, thick}
            \addlegendentry{Ours};
            \addlegendimage{gray,opacity=0.5,line width=2pt}
            \addlegendentry{Ground Truth};
            
            \addplot[color=gray,opacity=0.5,line width=2.5pt] table[x=T,y=Y]{\pathtoresults train_trajectory_D500_index0_karman.csv};
            \addplot[color=blue] table[x=T,y=Ymodel]{\pathtoresults train_trajectory_D500_index0_karman.csv};
            \addplot[color=gray,opacity=0.5,line width=2.5pt] table[x=T,y=Y]{\pathtoresults train_trajectory_D500_index13_karman.csv};
            \addplot[color=blue] table[x=T,y=Ymodel]{\pathtoresults train_trajectory_D500_index13_karman.csv};
            \addplot[color=gray,opacity=0.5,line width=2.5pt] table[x=T,y=Y]{\pathtoresults train_trajectory_D500_index25_karman.csv};
            \addplot[color=blue] table[x=T,y=Ymodel]{\pathtoresults train_trajectory_D500_index25_karman.csv};
            \addplot[color=gray,opacity=0.5,line width=2.5pt] table[x=T,y=Y]{\pathtoresults train_trajectory_D500_index3_karman.csv};
            \addplot[color=blue] table[x=T,y=Ymodel]{\pathtoresults train_trajectory_D500_index3_karman.csv};
            \addplot[color=gray,opacity=0.5,line width=2.5pt] table[x=T,y=Y]{\pathtoresults train_trajectory_D500_index4_karman.csv};
            \addplot[color=blue] table[x=T,y=Ymodel]{\pathtoresults train_trajectory_D500_index4_karman.csv};
        \end{axis}
        \begin{axis}[
                name=plot2,
                at=(plot1.east),
                xshift=0.1cm,
                anchor=west,
                height = 4cm,
                width=0.515\textwidth,
                xmin=0, xmax=1,
                xlabel={$t$},
                xmajorgrids,
                ymin=0.0075, ymax=.04,
                ylabel={$\|v\|$},
                ylabel near ticks, yticklabel pos=right,
                yticklabel style={xshift=0pt},
                ylabel style={yshift=-10pt},
                ymajorgrids,
                legend columns=3,
                legend style={column sep=1pt, text opacity=1,draw opacity=1,fill opacity=0.2,anchor=north west},
                legend pos=north west,
                title={Test Trajectories},
            ]
            \addlegendimage{blue, thick}
            \addlegendentry{Ours};
            \addlegendimage{gray,opacity=0.5,line width=2pt}
            \addlegendentry{Ground Truth};
            
            \addplot[color=gray,opacity=0.5,line width=2.5pt] table[x=T,y=Y]{\pathtoresults test_trajectory_D500_index0_karman.csv};
            \addplot[color=blue] table[x=T,y=Ymodel]{\pathtoresults test_trajectory_D500_index0_karman.csv};
            \addplot[color=gray,opacity=0.5,line width=2.5pt] table[x=T,y=Y]{\pathtoresults test_trajectory_D500_index1_karman.csv};
            \addplot[color=blue] table[x=T,y=Ymodel]{\pathtoresults test_trajectory_D500_index1_karman.csv};
            \addplot[color=gray,opacity=0.5,line width=2.5pt] table[x=T,y=Y]{\pathtoresults test_trajectory_D500_index2_karman.csv};
            \addplot[color=blue] table[x=T,y=Ymodel]{\pathtoresults test_trajectory_D500_index2_karman.csv};
            \addplot[color=gray,opacity=0.5,line width=2.5pt] table[x=T,y=Y]{\pathtoresults test_trajectory_D500_index3_karman.csv};
            \addplot[color=blue] table[x=T,y=Ymodel]{\pathtoresults test_trajectory_D500_index3_karman.csv};
            \addplot[color=gray,opacity=0.5,line width=2.5pt] table[x=T,y=Y]{\pathtoresults test_trajectory_D500_index4_karman.csv};
            \addplot[color=blue] table[x=T,y=Ymodel]{\pathtoresults test_trajectory_D500_index4_karman.csv};
        \end{axis}
    \end{tikzpicture}
\caption{
    Observable trajectories of the simulated cylinder flow and the surrogate model \textbf{Left:} Samples from the training data are depicted. \textbf{Right:} The test data is depicted.
    }\label{figure:KarmanTrajectories}
\end{figure}
}

%% file: plots/appendix_karman_experiment.tex
\begin{figure}
    \centering
    \begin{subfigure}[b]{0.495\textwidth}
        \centering
        \includegraphics[width=\textwidth]{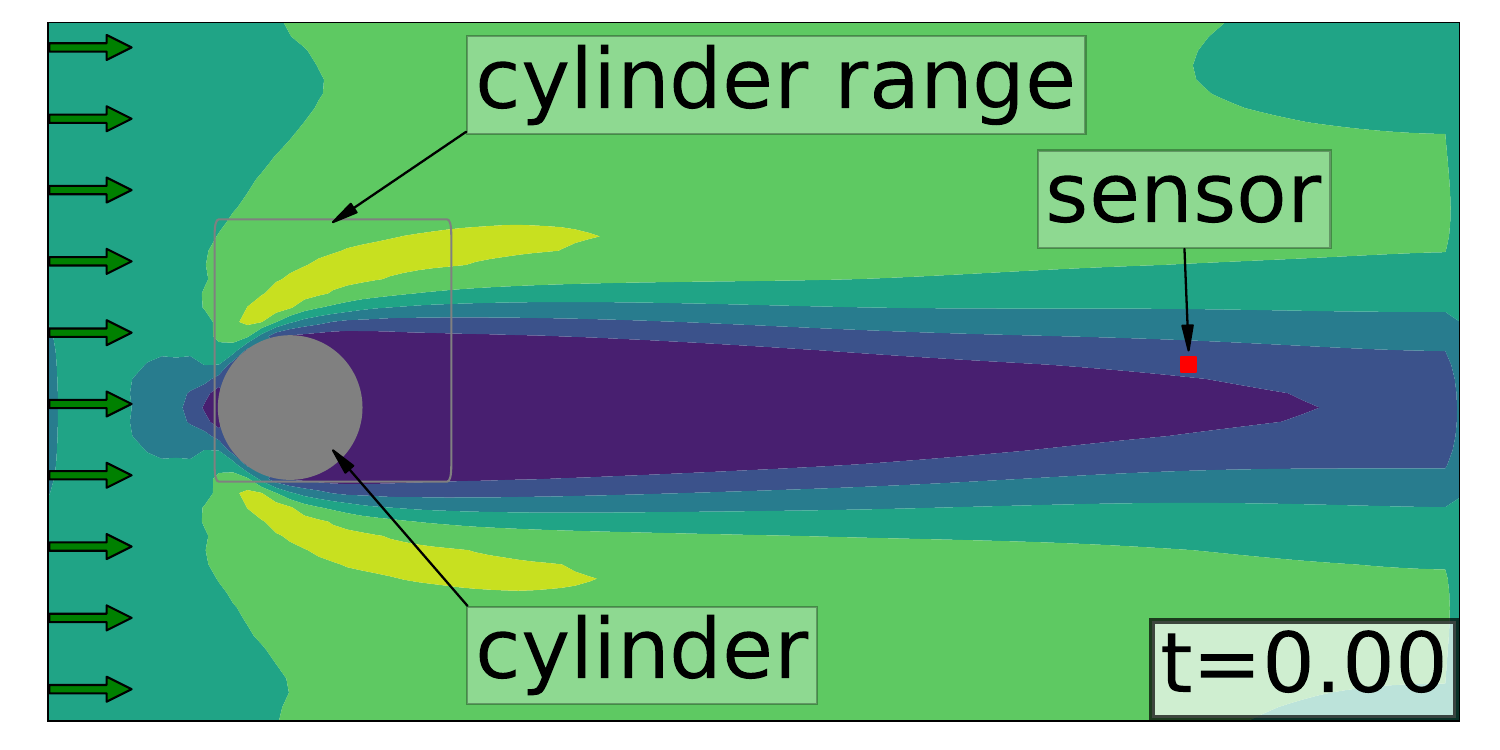}
    \end{subfigure}
    \hfill
    \begin{subfigure}[b]{0.495\textwidth}
        \centering
        \includegraphics[width=\textwidth]{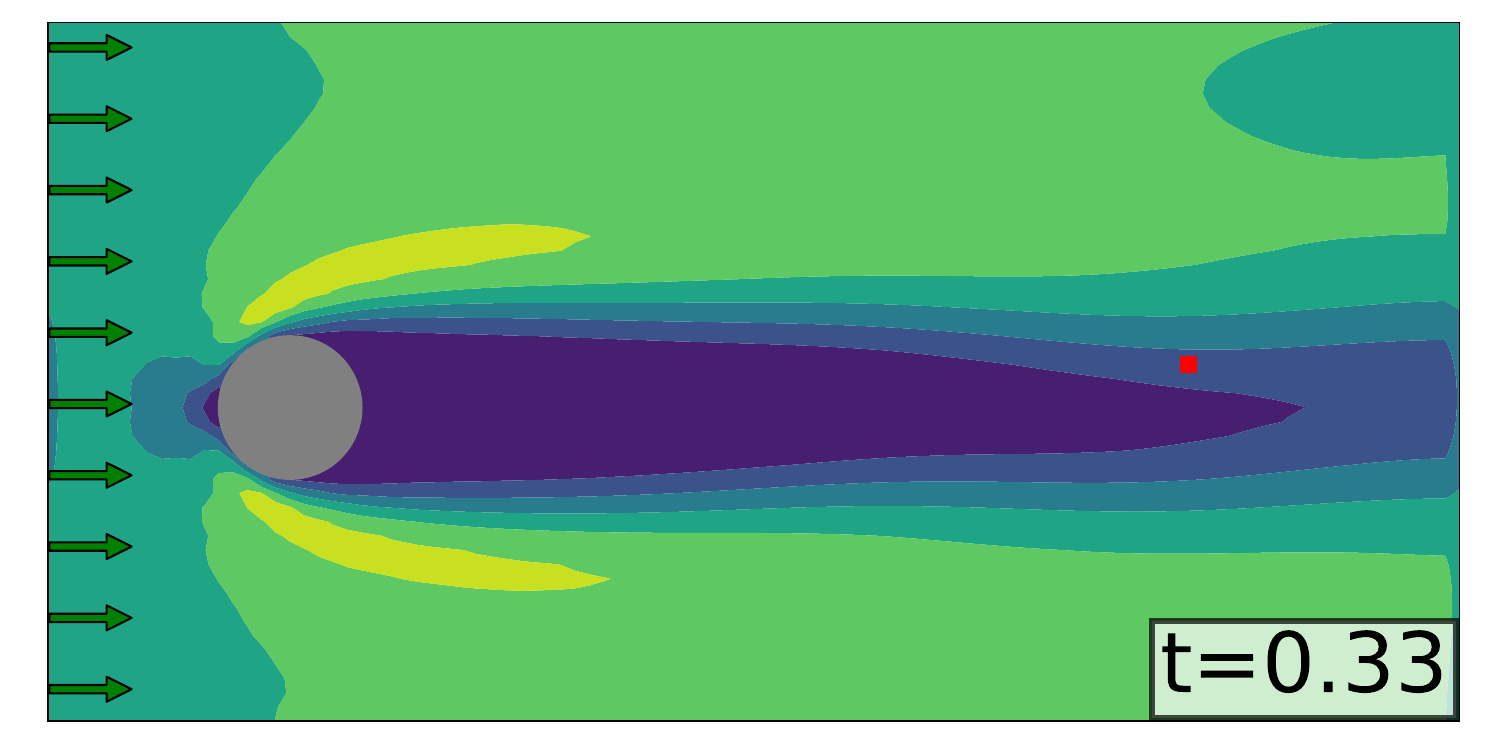}
    \end{subfigure}
    \vskip\baselineskip
    \begin{subfigure}[b]{0.495\textwidth}
        \centering
        \includegraphics[width=\textwidth]{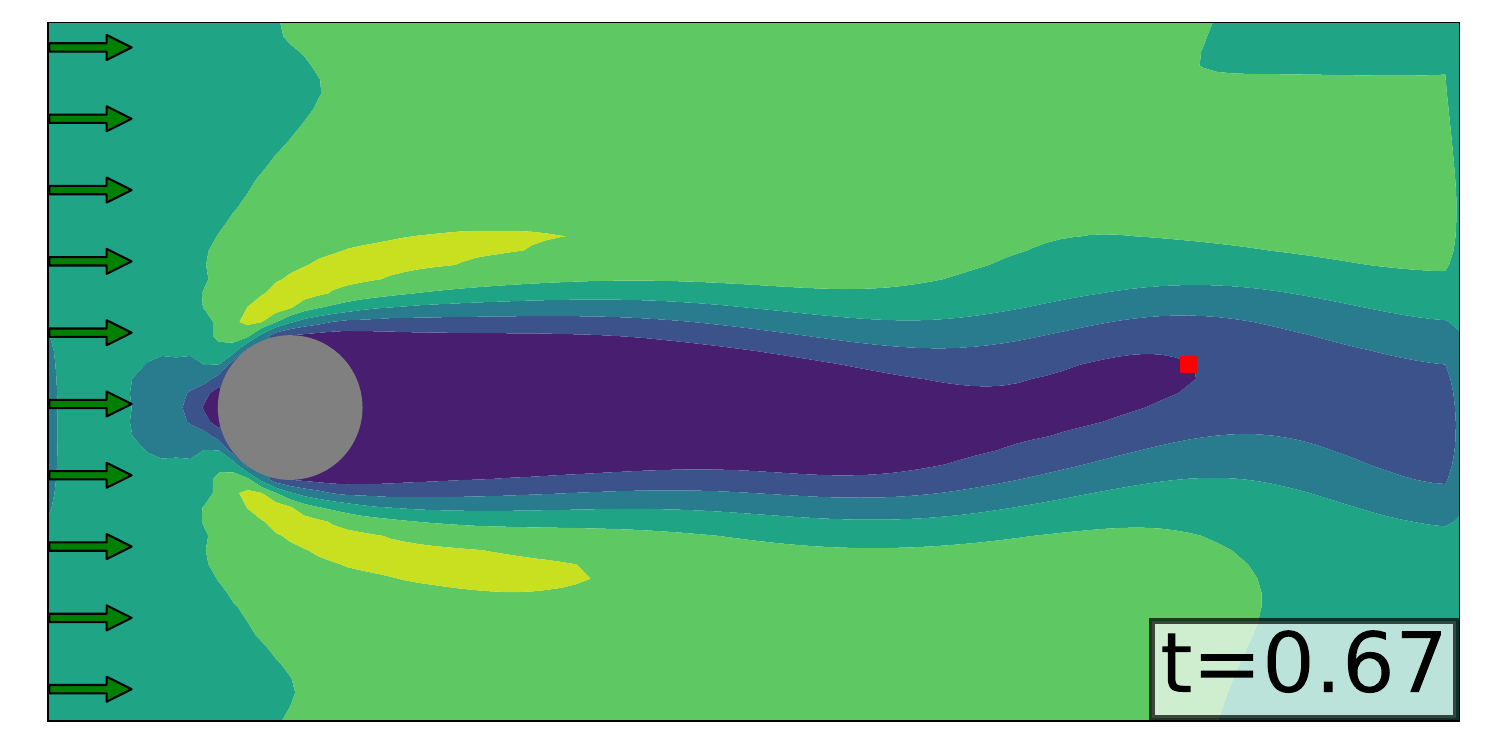}
    \end{subfigure}
    \hfill
    \begin{subfigure}[b]{0.495\textwidth}
        \centering
        \includegraphics[width=\textwidth]{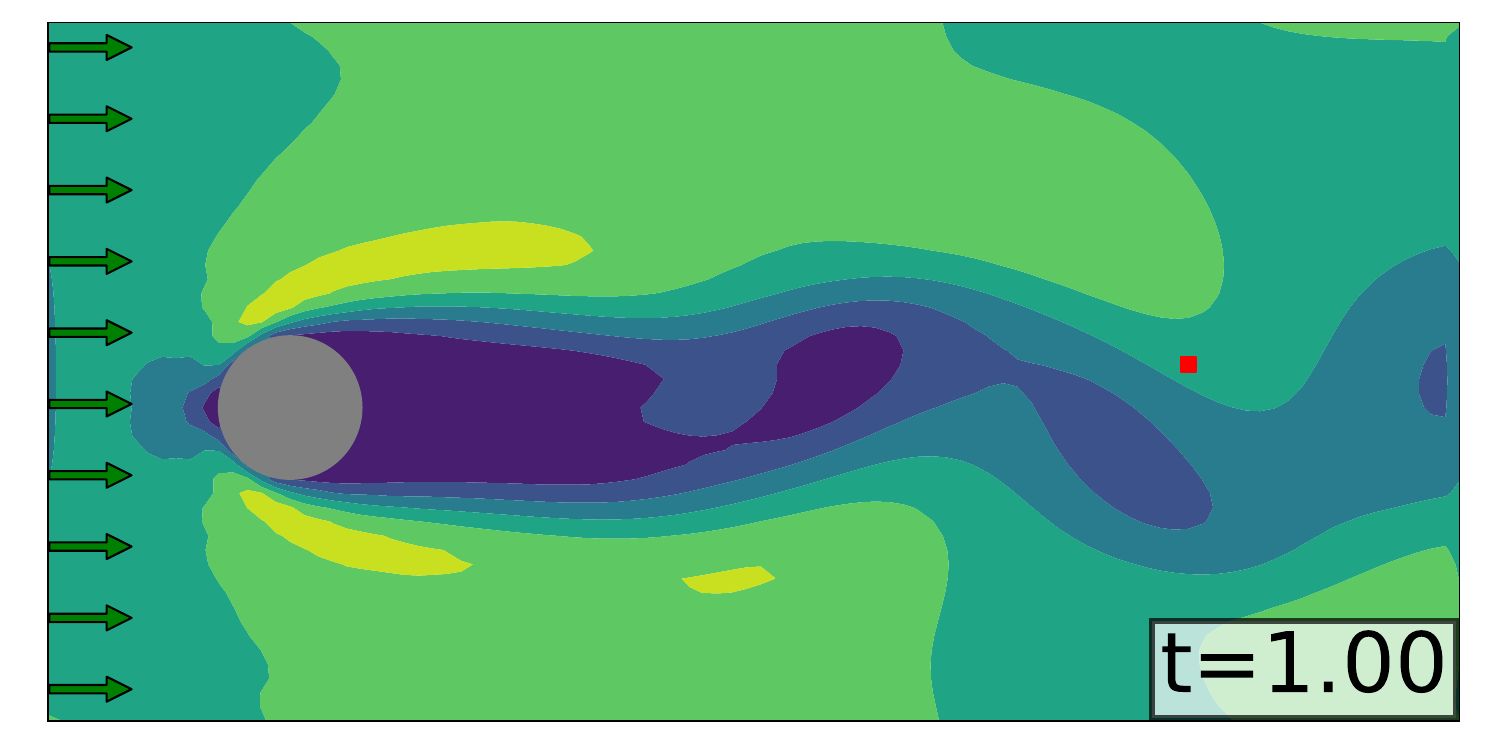}
    \end{subfigure}
    \caption{Velocity magnitudes in a developing Kármán vortex street behind a cylinder at different times. Yellow color indicates high and blue low magnitude.}
    \label{figure:cylinderFlow}
\end{figure}